\pdfoutput=1
\documentclass[10pt,twocolumn,letterpaper]{article}

\usepackage[pagenumbers]{cvpr} 

\usepackage{times}
\usepackage{epsfig}
\usepackage{graphicx}
\usepackage{amsmath}
\usepackage{amssymb}
\usepackage{bm}
\usepackage{dsfont}
\usepackage{enumitem}

\usepackage{color}
\usepackage[table]{xcolor}
\usepackage{soul}
\usepackage{booktabs}
\usepackage[normalem]{ulem}

\usepackage{pifont}
\newcommand{\cmark}{\ding{51}}%
\newif\ifdraft
\draftfalse

\usepackage[normalem]{ulem}

\ifdraft
 \newcommand{\WL}[1]{{\color{blue}{\bf WL: #1}}}
 
 \newcommand{\BT}[1]{{\color{green}{\bf BT: #1}}}
 
 \newcommand{\HC}[1]{{\color{purple}{\bf HC: #1}}}
 
 \newcommand{\VV}[1]{{\color{orange}{\bf VV: #1}}}
 
  \newcommand{\PF}[1]{{\color{red}{\bf PF: #1}}}
 
\else
 \newcommand{\WL}[1]{}
 
 \newcommand{\BT}[1]{}
 
 \newcommand{\HC}[1]{}
 
 \newcommand{\VV}[1]{}
 
 \newcommand{\PF}[1]{}
 
\fi

\newcommand{\comment}[1]{}
\newcommand{\parag}[1]{\paragraph{#1}}

\newcommand{\real}{\mathbb{R}}

\newcommand{\s}{\mathbf{s}}
\newcommand{\boldS}{\mathbf{S}}
\newcommand{\boldV}{\mathbf{V}}

\newcommand{\x}{\mathbf{x}}
\newcommand{\X}{\mathbf{X}}
\newcommand{\Y}{\mathbf{Y}}

\newcommand{\y}{\mathbf{y}}
\newcommand{\boldv}{\mathbf{v}}

\usepackage{multirow}

\usepackage[pagebackref,breaklinks,colorlinks]{hyperref}

\usepackage[capitalize]{cleveref}
\crefname{section}{Sec.}{Secs.}
\Crefname{section}{Section}{Sections}
\Crefname{table}{Table}{Tables}
\crefname{table}{Tab.}{Tabs.}

\def\cvprPaperID{XXXX} 
\def\confName{CVPR}
\def\confYear{2022}

\begin{document}

\title{Learning to Align Sequential Actions in the Wild}

\author{
	Weizhe Liu\textsuperscript{1}
	\quad
	Bugra Tekin\textsuperscript{2}
	\quad
     Huseyin Coskun\textsuperscript{3}
    \quad
    Vibhav Vineet\textsuperscript{2}
    \quad
	Pascal Fua\textsuperscript{1}
    \quad 
    Marc Pollefeys\textsuperscript{2,4} \\
	\textsuperscript{1} EPFL 
    \quad \textsuperscript{2} Microsoft  
    \quad \textsuperscript{3} Technische Universität München
    \quad \textsuperscript{4} ETH Zurich\\
}

\maketitle

\begin{abstract}

State-of-the-art methods for self-supervised sequential action alignment rely on deep networks that find correspondences across videos in time. They either learn frame-to-frame mapping across sequences, which does not leverage temporal information, or assume monotonic alignment between each video pair, which ignores variations in the order of actions. As such, these methods are not able to deal with common real-world scenarios that involve background frames or videos that contain non-monotonic sequence of actions.

In this paper, we propose an approach to align sequential actions in the wild that involve diverse temporal variations. To this end, we propose an approach to enforce temporal priors on the optimal transport matrix, which leverages temporal consistency, while allowing for variations in the order of actions. Our model accounts for both monotonic and non-monotonic sequences and handles background frames that should not be aligned. We demonstrate that our approach consistently outperforms the state-of-the-art in self-supervised sequential action representation learning on four different benchmark datasets. 

\end{abstract}
\section{Introduction}
\label{sec:intro}

Understanding human activities in video sequences is important for applications such as human-computer interaction, video analysis, robot learning, and surveillance.  In recent years, a significant amount of research has focused on supervised, coarse-scale action understanding. Most of the work focuses on predicting explicit classes for clips corresponding to a certain limited set of action categories in a supervised fashion \cite{Carreira17,Tran15,Tran18,Wang18,Coskun21,Kwon21}. While giving a categorical understanding of human behavior, such techniques do not provide a fine-grained analysis of human action. 
Furthermore, the dependence on per-frame labels requires a large amount of human effort that does not scale up to many different types of subjects, environments, and scenarios.
For such supervised methods, it is also not always clear what exhaustive set of labels is required for a fine-grained understanding of videos.
Thus, recent papers~\cite{Dwibedi19,Haresh21} advocate self-supervised learning of video representation without frame-wise action labels. They rely on the fact that human activities often involve many sequential steps in a predictable order. To drink water, one might grab a mug, drink, and then put the mug down. To change a tire, one would  first lift the vehicle off the ground, remove the wheel, and replace it by a spare one. Assuming the order is set, visual representations can be learned from multiple videos of the same activity by temporal alignment of the frames. 

\begin{figure}[t]
\centering
\begin{tabular}{cc}
\includegraphics[width=.45\linewidth]{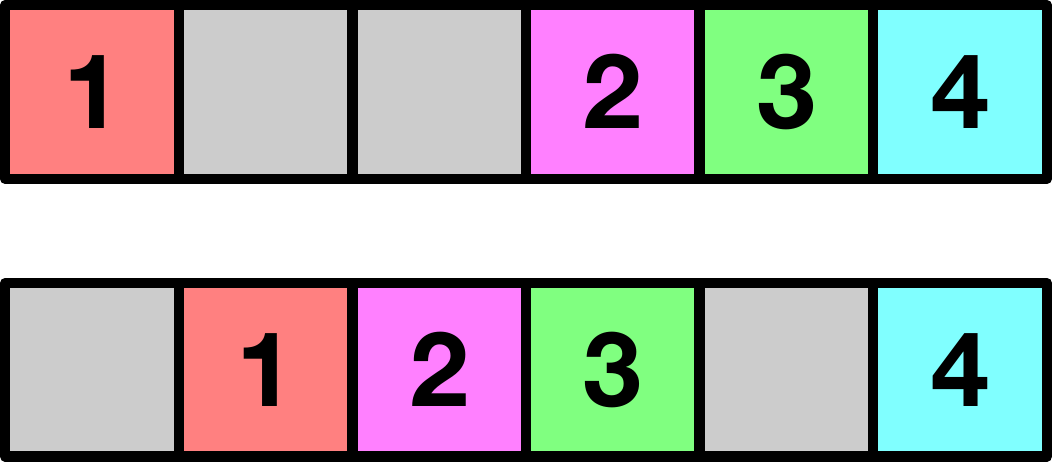}&
\includegraphics[width=.45\linewidth]{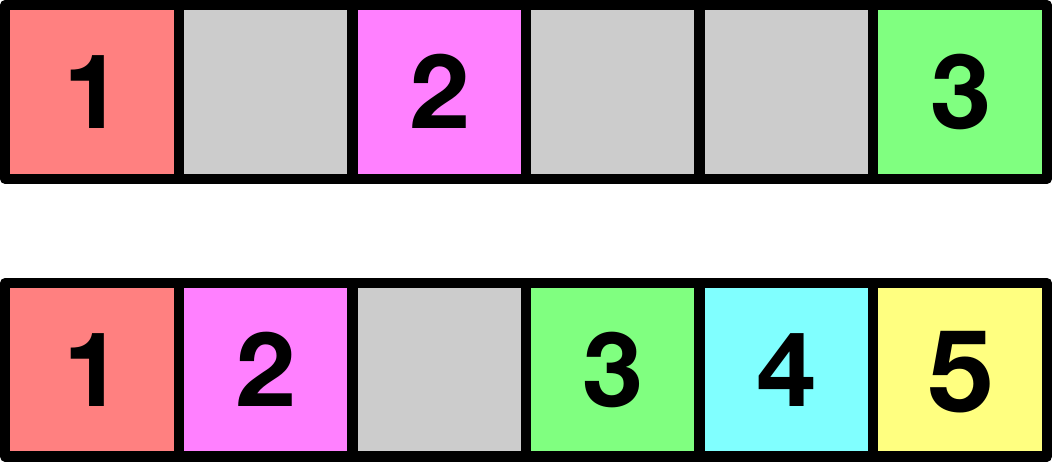}\\
\small{\bf{(a)}}&
\small{\bf{(b)}}\\
\multicolumn{2}{c}{
\includegraphics[width=.7\linewidth]{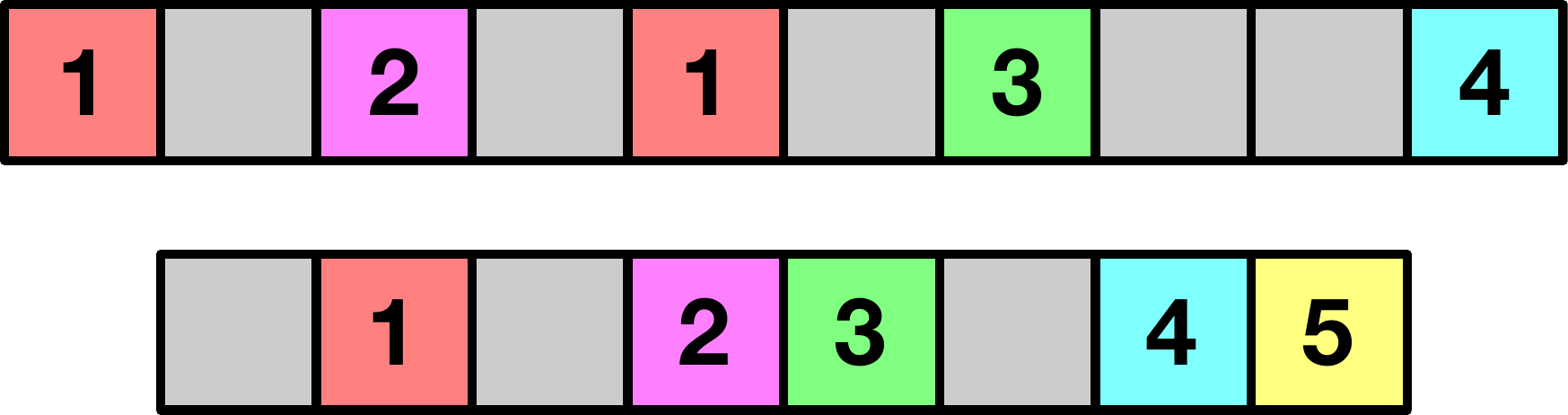}}\\

\multicolumn{2}{c}{\small{\bf{(c)}}}
\end{tabular}
\caption{\small {\bf Temporal Variations~\cite{Dogan18}.} {\bf(a)} \emph{Background frames}, depicted as gray blocks, are not related to the major activity. {\bf(b)} Frames with number $4$ and $5$ are \emph{redundant frames} which only exist in one sequence but not in the other. {\bf(c)} Frames with action $1$ in the first sequence occur before and after action $2$ and forms a sequence of \emph{non-monotonic frames}. Our approach explicitly tackles all of these temporal variations and is suitable for aligning sequential actions in a broad context.
\vspace{-3mm}
}
\label{fig:intro}
\end{figure}

\begin{figure}[t]
\centering
\begin{tabular}{c}
\includegraphics[width=0.8\linewidth]{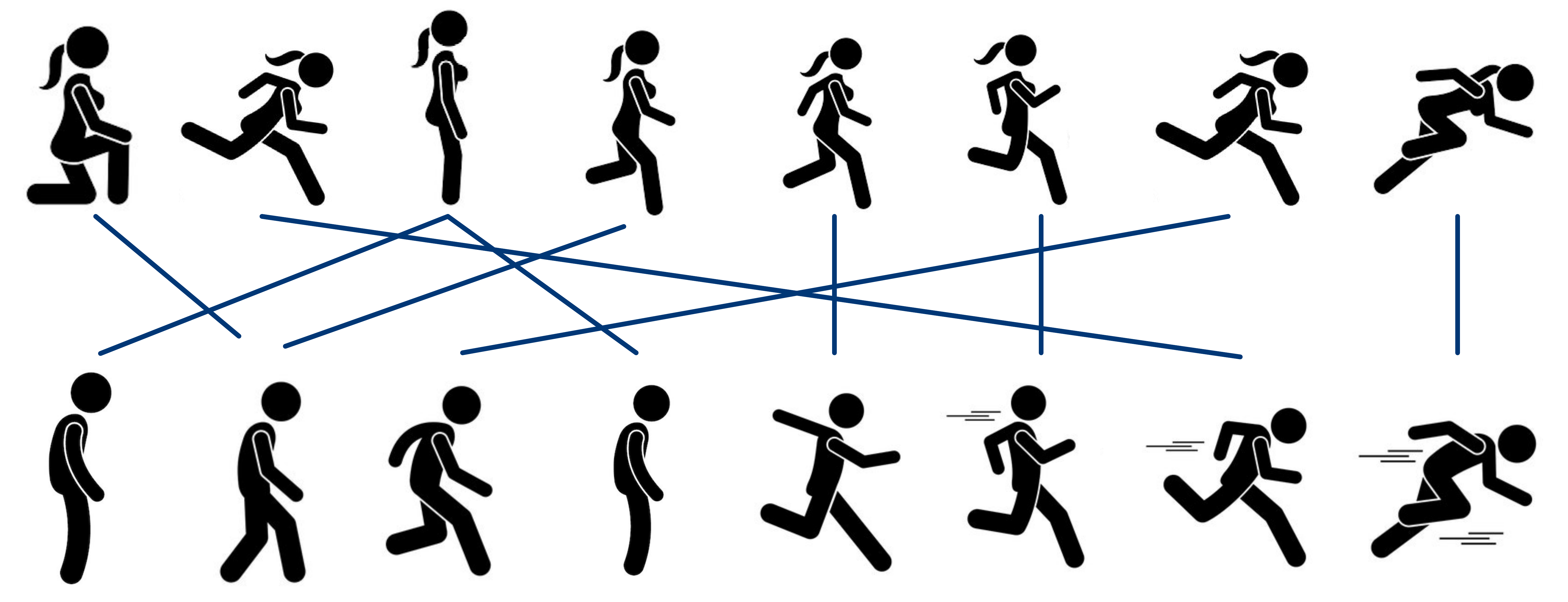}\\
\vspace{2mm}
\small{\bf{(a)} Alignment without Temporal Prior }\\
\includegraphics[width=0.8\linewidth]{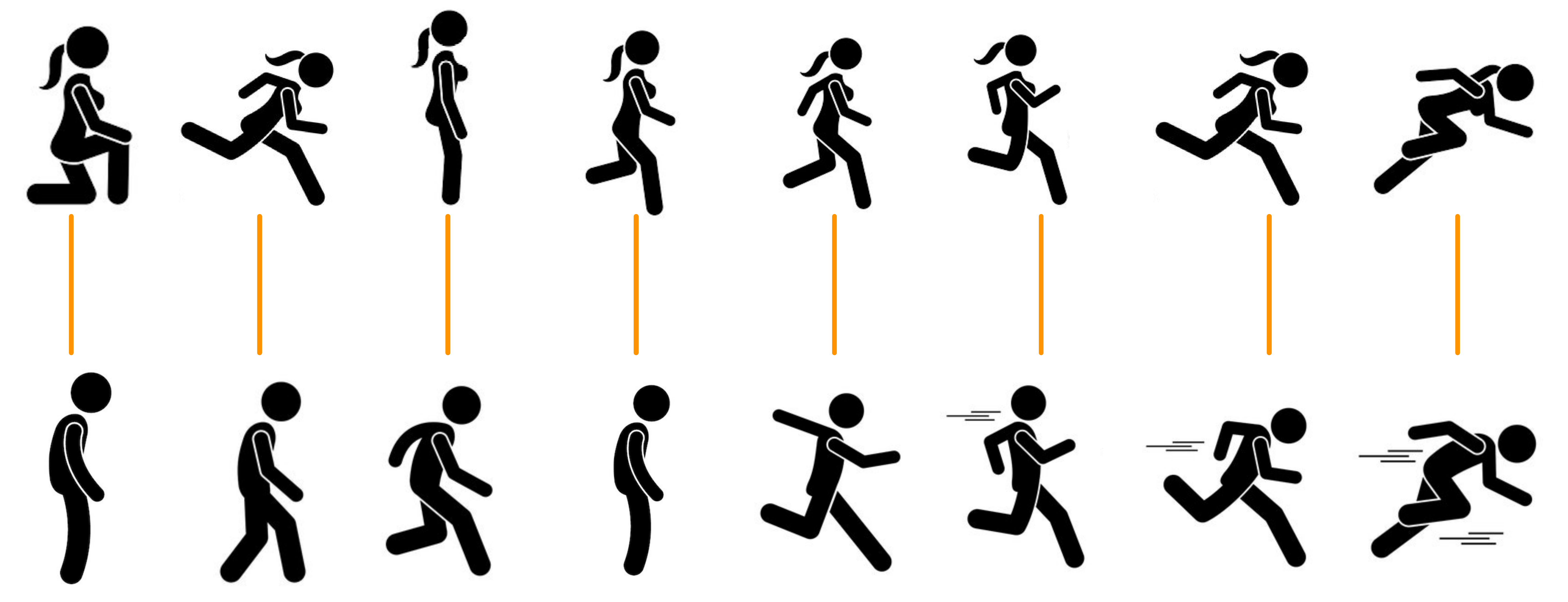}\\
\vspace{2mm}
\small{\bf{(b)} Monotonic Alignment}\\
\includegraphics[width=0.8\linewidth]{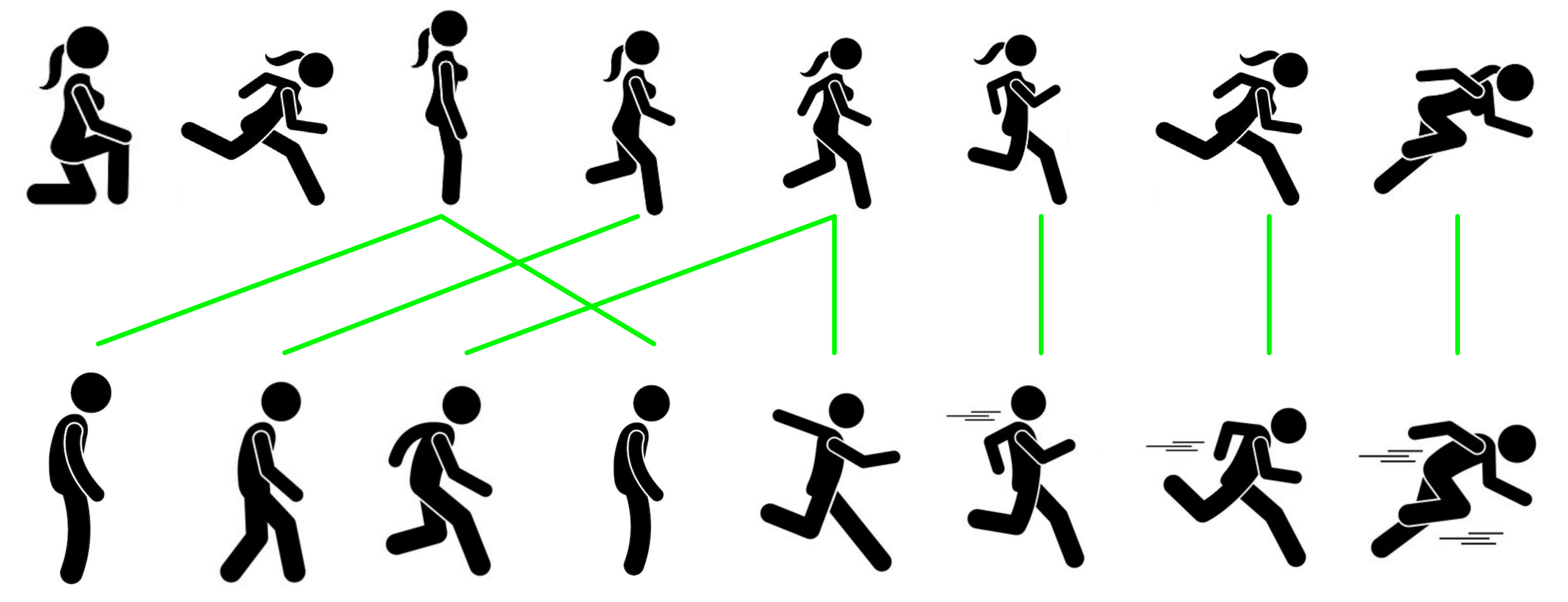}\\
\vspace{2mm}
\small{\bf{(c)} VAVA}\\
\end{tabular}
\vspace{-5mm}
\caption{\small {\bf Types of Alignment Priors.} Two example sequences of people running. The first sequence shows a person starting from a professional crouching position while the second sequence shows another person starting from a normal standing position.  {\bf (a)} Alignment without temporal prior is based on pure appearance similarity, therefore the starting ``crouching'' action of the first sequence is incorrectly aligned with the ``speeding up'' action in the second one. {\bf (b)} Monotonic alignment enforces pure monotonic order, therefore even though two actions look quite different, they can be incorrectly aligned. 
{\bf (c)} By contrast, our approach, VAVA,  enforces temporal priors to address non-monotonic frames and gracefully handles unmatched frames (\eg crouching position in the first sequence), resulting in accurate alignment between sequences. 
\vspace{-5mm}
}
\label{fig:compare}
\end{figure}

This is often done by monotonically aligning the frames\cite{Hadji21}, which assumes that actions always occur in the same order. However, in most real-world sequences, this is not the case and temporal deviations such as those depicted by Fig.~\ref{fig:intro} do occur. They can be summarized as follows:
\vspace{-1mm}
\begin{itemize}[leftmargin=4.5mm]
    \item \emph{Background frames}: Frames that are not related to the major activity and should therefore not be aligned. For example, you might get a phone call while changing the tire. In this case, the ``phone call'' frames are background frames that are not related to the major activity and should be ignored. 
    \vspace{-1mm}
    \item \emph{Redundant frames}: Frames that only exist in one sequence but not in the other. For example, one person might put on gloves before changing the tire while another does not.  In this case, the ``glove wearing'' frames are redundant and should be ignored as well.
    \vspace{-1mm}
    \item \emph{Non-monotonic frames}: Frames that occur in non-monotonic order. For example, while changing the tire, you lift your vehicle off the ground and try to remove the tire, only to realize that you have not lifted high enough. You then go back to the previous action, that is, lifting, before proceeding with the remaining actions. 
    \vspace{-1mm}
\end{itemize}
Our method aims at tackling all these cases and reduces the stringent assumptions of earlier work on the temporal sequence of actions. For this purpose, we propose an approach for learning temporal correspondences across videos through a novel alignment framework. Our model accounts for temporal variations exhibited across real-world sequences with a differentiable deep network formulation that relies on an optimal transport loss. While optimal transport is able to align non-monotonic sequences based on frame-wise matching of the features computed from individual frames, it ignores temporal smoothness and ordering relationships of the videos. To remedy this, we introduce temporal priors on the transportation matrix that the optimal transport algorithm takes as input. This accounts for the temporal structure of the sequence and enforces time consistency during alignment in a flexible way.  This is unlike previous work that either ignores temporal priors within  sequences~\cite{Dwibedi19} or enforces monotonic alignment between pairs of videos~\cite{Haresh21}, as depicted by Fig.~\ref{fig:compare}.

In particular, we enforce a temporal prior by modeling the diagonal of the optimal transport matrix with an adaptive Gaussian Mixture Model (GMM). Our temporal prior effectively favors transportation of one sequence to the elements in the nearby temporal positions of the other sequence, and, hence respects the overall temporal structure and order of the sequences during alignment. At the same time, our optimal transport based formulation aims to find ideal frame-wise matches and handles non-monotonic frames. To explicitly handle background and redundant frames, we further propose an approach, which introduces an additional virtual frame in the optimal transport matrix so that unmatched frames are explicitly assigned to it. Furthermore, since enforcing temporal priors on video alignment generally suffers from converging to trivial solutions~\cite{Hadji21}, we introduce a novel inter-video contrastive loss to regularize the learning process. In particular, our contrastive loss optimizes for disentangled video representations, i.e., videos that are close in terms of their similarity given by optimal transport are mapped to spatially nearby points in the embedding space and vice versa.

Our contributions can be summarized as follows: First, we propose a self-supervised learning approach that aligns sequential actions in-the-wild, which feature a diverse set of temporal variations. Second, we enforce adaptive temporal priors on optimal transport, which could efficiently handle non-monotonic frames while respecting the local temporal structure of sequences. Third, we extend the optimal transport formulation with an additional virtual frame that actively handles redundant and background frames that should not be matched. Finally, to prevent our model from converging to trivial solutions, we propose a novel contrastive loss term that regularize the learning of optimal transport matrix. In Sec.~\ref{sec:experiments}, we show quantitatively that these contributions allow us to reliably learn robust temporal correspondences and align sequential actions in real-world settings. Our self-supervised approach, which we call \emph{Variation-Aware Video Alignment (VAVA)}, uses temporal alignment as a pretext to learn visual representations that are effective in downstream tasks, such as action phase classification and tracking the progress of an action, and significantly outperforms state-of-the-art methods on four different benchmark datasets.
\section{Related Work}
\parag{Self-Supervised Video Representation Learning.}\label{sec:related_RL} Temporal
information in videos provide rich supervision signal to learn strong spatio-temporal representations~\cite{Diba21,Qian21,Feichtenhofer21}. This contrasts to single-image based approaches~\cite{Feng19,Gidaris18,Larsson16,Larsson17,Noroozi17,Hinton94,Chen20,He20,Yuan21,Liu21,Kotar21,Dwibedi21,Zheng21} that only rely on spatial signal. Misra~\emph{et al.}~\cite{Misra16} introduce the idea of learning such visual representations by estimating the order of shuffled video frames. Inspired by the success of this approach, several recent papers focused on designing a novel pretext task using temporal information, such as predicting future frames~\cite{Diba19,Srivastava15,Vondrick16} or their embeddings~\cite{Gammulle19,Han19}; estimating the order of frames~\cite{Misra16, Choi20,Lee17,Fernando17,Xu19} or the direction of video~\cite{Wei18}. Another line of research focuses on using temporal coherence~\cite{Goroshin15,Zou11,Zou12,Mobahi09,Hadsell06,Bengio09} as supervision signal.

However, these methods usually optimize over a single video at a time, therefore they exploit less information compared with approaches that jointly optimize over a pair of videos~\cite{Dwibedi19,Haresh21}. Furthermore, such visual representations are learned by maximizing the similarity of two randomly cropped and augmented clips from the same video~\cite{Pan21, Qian21,Behrmann21,Feichtenhofer21,Lin21}. This requires training videos that contain the exact same single action. However, in real world scenarios, a complex human activity typically involves multiple actions and even background frames. Another limitation of these approaches is that they aim to learn coarse clip-wise visual representations, therefore they are not suitable for frame-wise downstream tasks like fine-grained action recognition. In contrast to these methods, we propose a self-supervised learning strategy that can learn frame-wise representations from unconstrained videos that involve sequential actions. 

\vspace{-5mm}
\parag{Video Alignment} \hspace{-3mm} is rather straightforward to address if the videos are synchronized. This can be done by using existing methods such as CCA~\cite{Anderson58,Andrew13} and DTW~\cite{Berndt94}. Recent trend in computer vision~\cite{Sermanet18} leverages deep networks and proposes to align videos by learning self-supervised visual representations from videos with the same human activity. In this regard, Sermanet~\emph{et al.}~\cite{Sermanet18} propose to learn cross-sequence visual representation by aligning synchronized multi-view videos that record exactly the same human actions from different viewpoints. As synchronized multi-view videos are not always available, this approach cannot be generalized to unconstrained settings.  Dwibedi~\emph{et al.}~\cite{Dwibedi19} address this issue by finding frame correspondences across unsynchronized videos with cycle consistency loss, however, this approach only looks for local matches across sequences and does not explicitly account for the global temporal structure of the videos.

Maybe the most similar works to our approach are~\cite{Haresh21,Hadji21}, which align video pairs with the assumption of strictly monotonic temporal order. As we explained in the introduction, this assumption is too strong and seldom happens naturally in real-world scenarios. In contrast to these methods, our approach does not require synchronized videos and learns to align video sequences from in-the-wild settings, which includes temporal variations, such as background frames, redundant frames and non-monotonic frames. As shown in Sec.~\ref{sec:experiments}, our approach consistently outperforms above methods and the margin is even larger if there were temporal variations. 

\vspace{-5mm}
\parag{Optimal Transport.} Optimal transport measures the dissimilarity between two probability distributions over a metric space. Given feature vectors associated to each entity and matrix of distances between them, it provides a way to establish correspondences between features that minimize the sum of distances. Besides, it also provides guarantees of optimality, separability, and completeness. These desirable properties have been leveraged for many different tasks, such as scene flow estimation~\cite{Puy20,Li21}, object detection~\cite{Ge21}, domain adaptation~\cite{Xu20}, classification~\cite{Serrurier21} and point matching~\cite{Sarlin20} that matches features in spatial domain. However, none of them focuses on sequence alignment as we do. One potential reason is that vanilla optimal transport formulation does not account for temporal priors, therefore the alignment is less reliable in the time domain, as depicted by Fig.~\ref{fig:compare}(a). One exception is~\cite{Su17}, which uses optimal transport only to measure the distance between skeleton sequences and does not learn a visual representation as we do. Besides, it only enforces monotonic temporal priors without accounting for the cases of temporal variations, therefore is less flexible than our approach which specifically addresses such situations. 
\section{Approach}

In this section, we first formalize the problem of self-supervised representation learning by aligning frames from pairs of video sequences (Sec.~\ref{sec:formulation}). After that, we present our approach for incorporating temporal priors in optimal transport to leverage temporal information and  handle non-monotonic frames  (Sec.~\ref{sec:temporal}). We then propose an effective way to deal with background and redundant frames (Sec.~\ref{sec:virtual}). 
Finally, we provide a summary of our loss function and model details (Sec.~\ref{sec:loss}). 

\subsection{Alignment by Optimal Transport}
\label{sec:formulation}

\begin{figure}[t]
\centering
\includegraphics[width=0.85\linewidth]{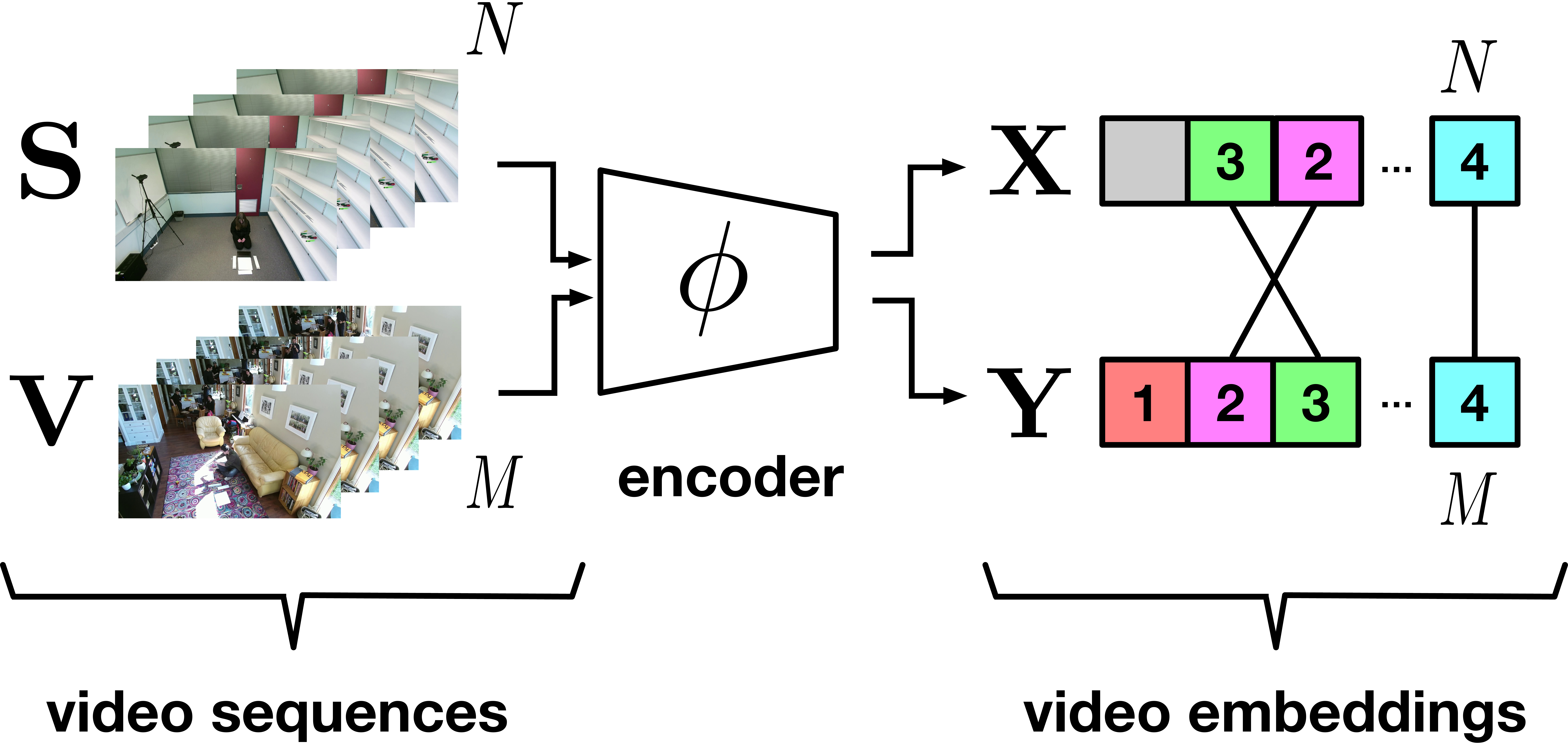}
\caption{\small {\bf Encoder Network and Video Embeddings.}}
\vspace{-4mm}
\label{fig:model}
\end{figure}

Given two sequences of video frames ${\boldS} =[\s_{1}, \s_{2}, ..., \s_{N} ]$ and ${\boldV} = [\boldv_{1}, \boldv_{2}, ..., \boldv_{M}]$, we take their respective embeddings to be
$\X = [\x_{1}, \x_{2}, ..., \x_{N} ]$ and $\Y = [\y_{1}, \y_{2}, ..., \y_{M}]$. $\X$ and $\Y$ are computed with an encoder network $\phi$, as depicted in Fig.~\ref{fig:model}.

If frames $\s_{i}$ and $\boldv_{j}$ represent the same fine-grained action, the distance between their respective embeddings, $\x_{i}$ and $\y_{j}$, should be small, otherwise, the distance should be large. Given such embeddings, Optimal Transport (OT) can be used to align two such sequences by first computing an $N \times M$ distance matrix, $\bm{D}$, whose components are Euclidean distances between embedding vectors, that is, $d(\x_{i},\y_{j})=\|\x_{i}-\y_{j}\|$. The optimal assignment, $d_{O}(\X,\Y)$, between the embeddings can be found by solving the following optimization problem:
\begin{align}
    d_{O}(\X,\Y) := \min_{\bm{T}\in U(\bm{\alpha},\bm{\beta})}<\bm{T},\bm{D}> \;
    \label{eq:orig_ot}   
\end{align}
Here, $<\cdot,\cdot>$ is the Frobenius dot product, and, $\bm{\alpha} = (\alpha_{1},...,\alpha_{N})$ and $\bm{\beta} = (\beta_{1},...,\beta_{M})$ are non-negative weights that sum to one and denote the relative importance of individual frames. As we have no reason to weigh one frame more than the others, we take $\alpha_{i}=1/N$ and $\beta_{j}=1/M$, for all $i$ and $j$. The set of all feasible transport matrices is represented with $U$. A valid transportation matrix in $U$ satisfies that the row and column-wise sum are equal to $\bm{\alpha}$ and $\bm{\beta}$~\cite{Cuturi13}, in particular:
\begin{align}
     U(\bm{\alpha},\bm{\beta}):=\{\bm{T} \in \real^{N \times M} |\bm{T}\bm{1}_{M}=\bm{\alpha},\bm{T}^{\top}\bm{1}_{N}=\bm{\beta}\}\;                         \label{eq:u}   
\end{align}
Eq.~\ref{eq:orig_ot} can be solved with linear programming, however, this is a computationally expensive procedure and is not suitable for training purposes. To address this issue, Cuturi~\cite{Cuturi13} proposes to regularize OT problem with an additional entropy term and solves it using Sinkhorn algorithm.
\begin{align}
    d_{O}(\X,\Y) := \min_{\bm{T}\in U(\bm{\alpha},\bm{\beta})}<\bm{T},\bm{D}> - \upsilon h(\bm{D}) \;     \label{eq:do}   
\end{align}
where $h$ is an entropy term that regularizes the problem and $\upsilon$ is a small scalar coefficient. 
Here, the entries of the transport matrix, \ie the $t_{i,j}$ coefficients of $\bm{T}$, can be understood to be proportional to the probability that frame $i$ in ${\boldS}$ is aligned with frame $j$ in ${\boldV}$. A large value of distance $d_{i,j}$ would correspond to a small value of $t_{i,j}$, which implies that these two frames are dissimilar and thus have a low chance of alignment. The benefit of such formulation is that we can enforce temporal priors by modeling $\bm{T}$ to follow a predefined temporal distribution.

\subsection{Enforcing Temporal Priors}
\label{sec:temporal}


\begin{figure}[t]
\centering
\begin{tabular}{cccc}
\includegraphics[width=.21\columnwidth]{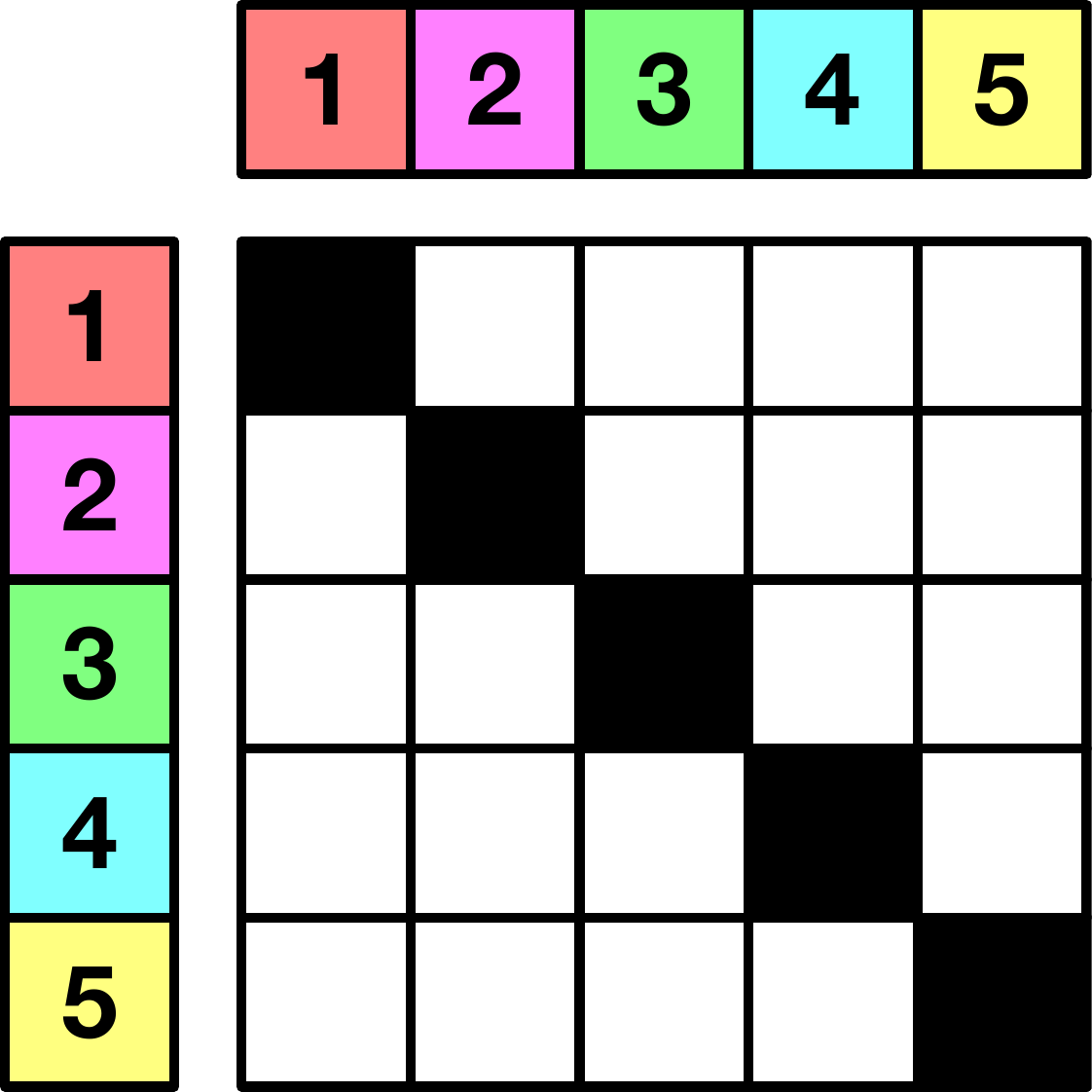}&
\includegraphics[width=.21\columnwidth]{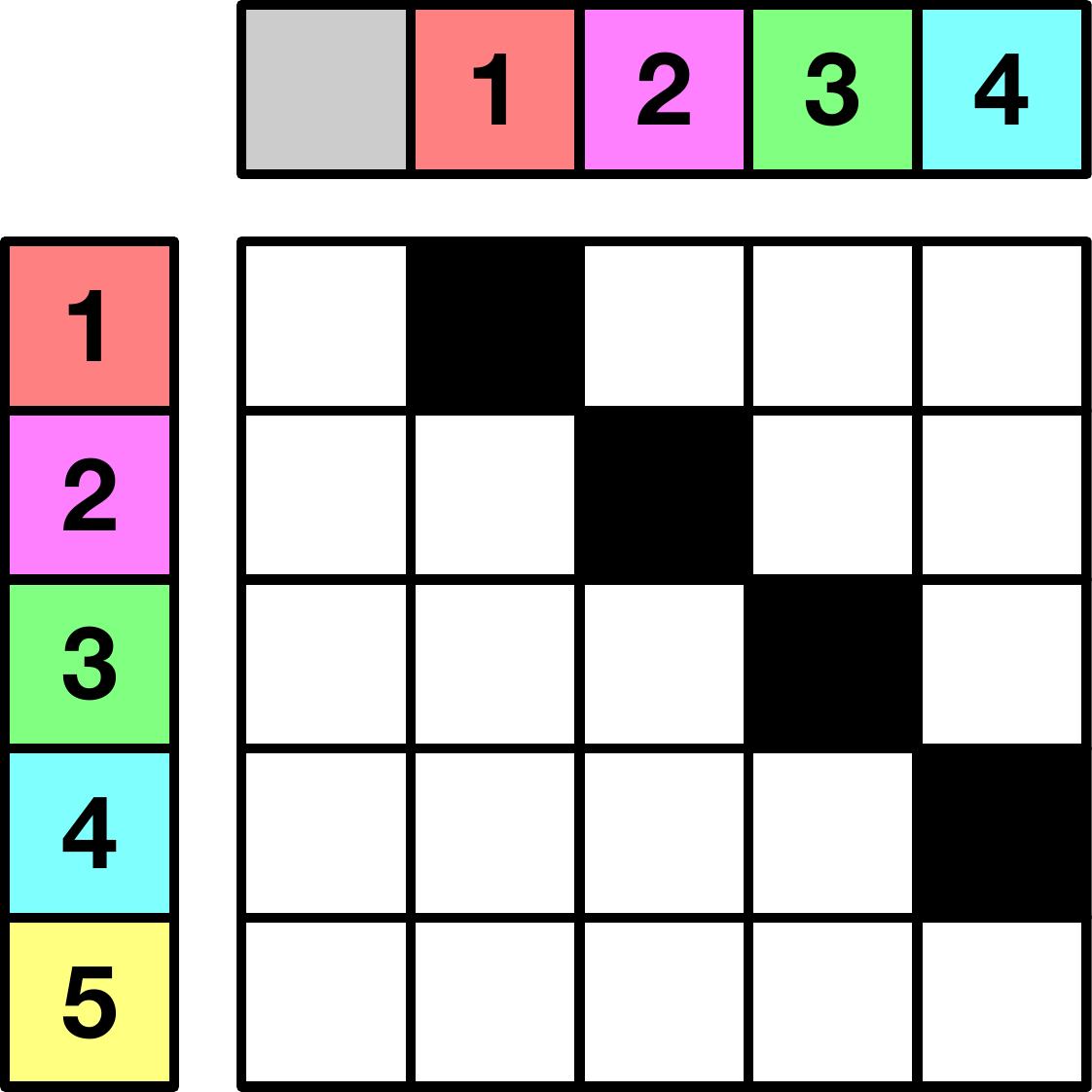}&
\includegraphics[width=.21\columnwidth]{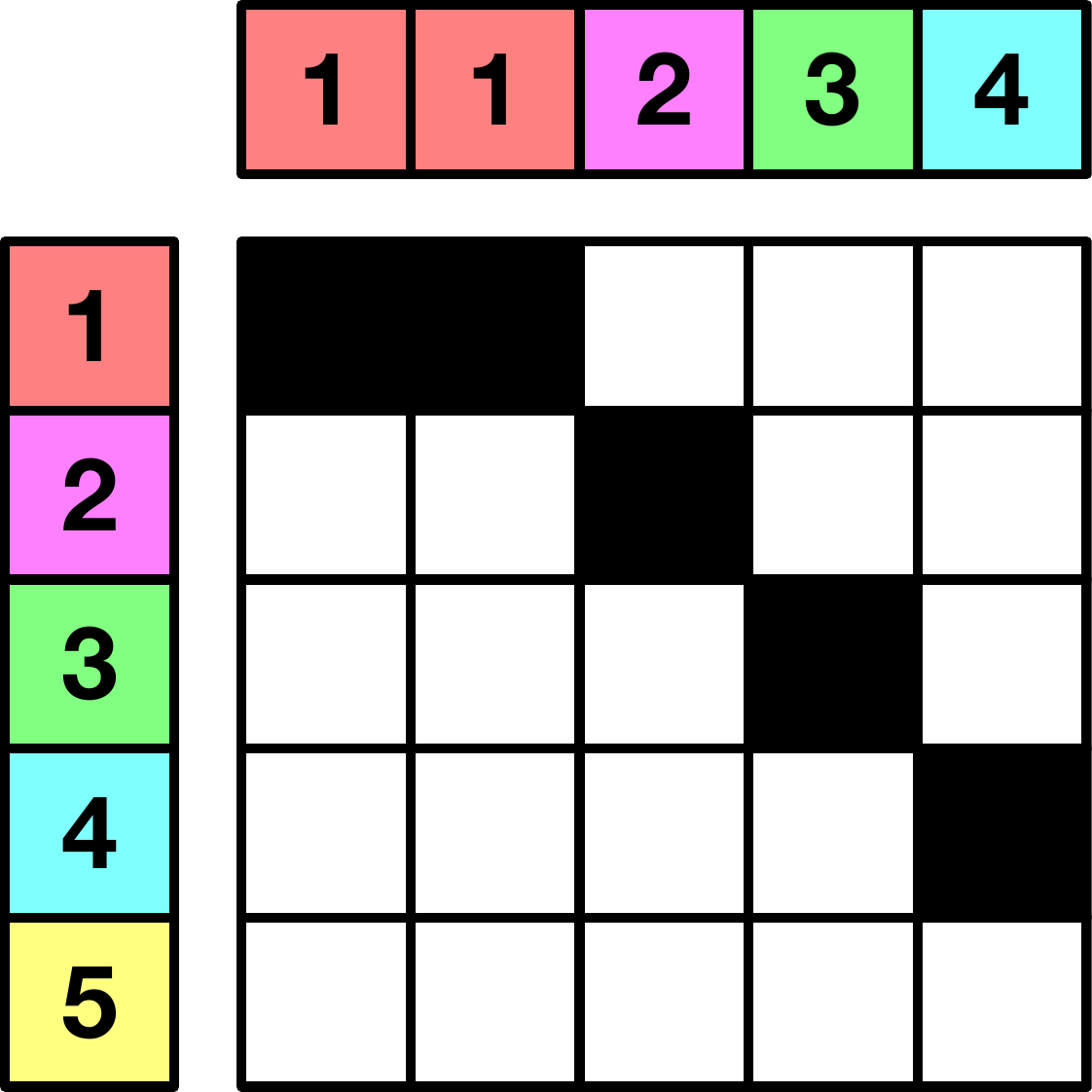}&
\includegraphics[width=.21\columnwidth]{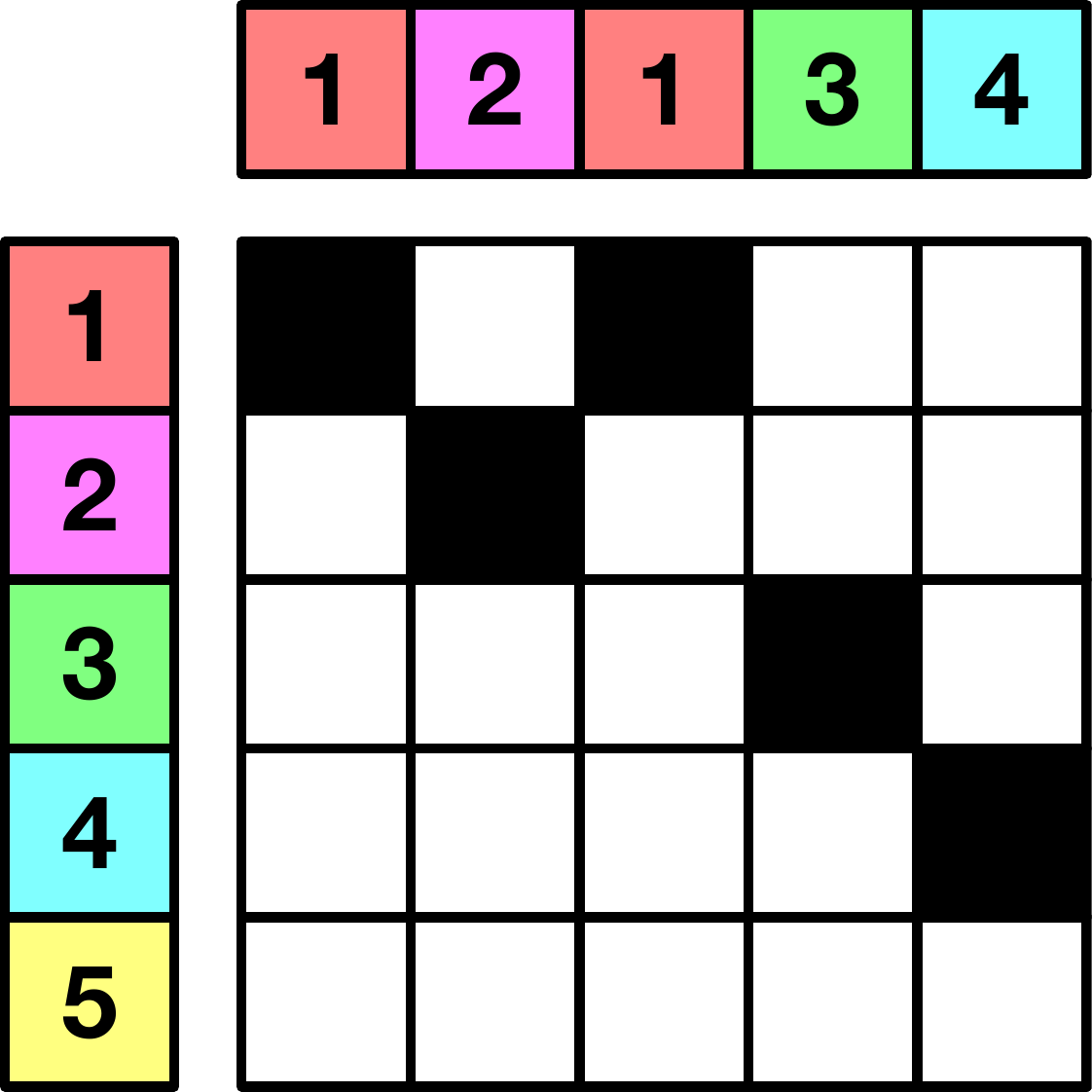}\\
\small{\bf{(a)}} &\small{\bf{(b)}} &\small{\bf{(c)}} &\small{\bf{(d)}}\\
\vspace{-6mm}
\end{tabular}
  \caption{ {\bf Assignment Variations.} (a) Two videos strictly follow the same temporal order, the assignment matrix has peak values along the diagonal. (b)  The activity from one video starts a bit earlier than the other one, hence the assignment matrix has peak values parallel to the diagonal.  (c) The action of one video is slower than the other, thus the assignment matrix has peak values that are near the diagonal without being strictly parallel to it. (d) Actions follow monotonic order in one sequence but not in the other.
  \vspace{-4mm}
 }
  \label{fig:variation}
  \end{figure}

While optimal transport measures the minimum cost of aligning two sequences, it completely ignores temporal ordering relationships and therefore does not exploit the temporal consistency that we know to be present in video sequences. In most cases, given multiple videos of the same activity, the temporal position of one sequence should only be aligned to elements in the nearby temporal positions of the other sequence. In an extreme case where the two sequences are perfectly aligned, the transport matrix $\bm{T}$ should be diagonal. In practice, this is far too strong a constraint. As depicted by Fig.~\ref{fig:variation}, the activity may start a bit earlier in one sequence than the other; it may be faster; the actions in one of the video sequences may be monotonic while the other ones are not. 

To capture temporal variations across sequences, while being able to optimally align two videos, we propose to enforce temporal priors on the optimal transport problem. To this end, we propose a novel prior distribution of transport matrix with an adaptive Gaussian Mixture Model (GMM) that comprises of two temporal priors. 

\begin{figure}[t]
  \centering
  \begin{tabular}{cccc}
  \hspace{-3mm}\includegraphics[width=.22\linewidth]{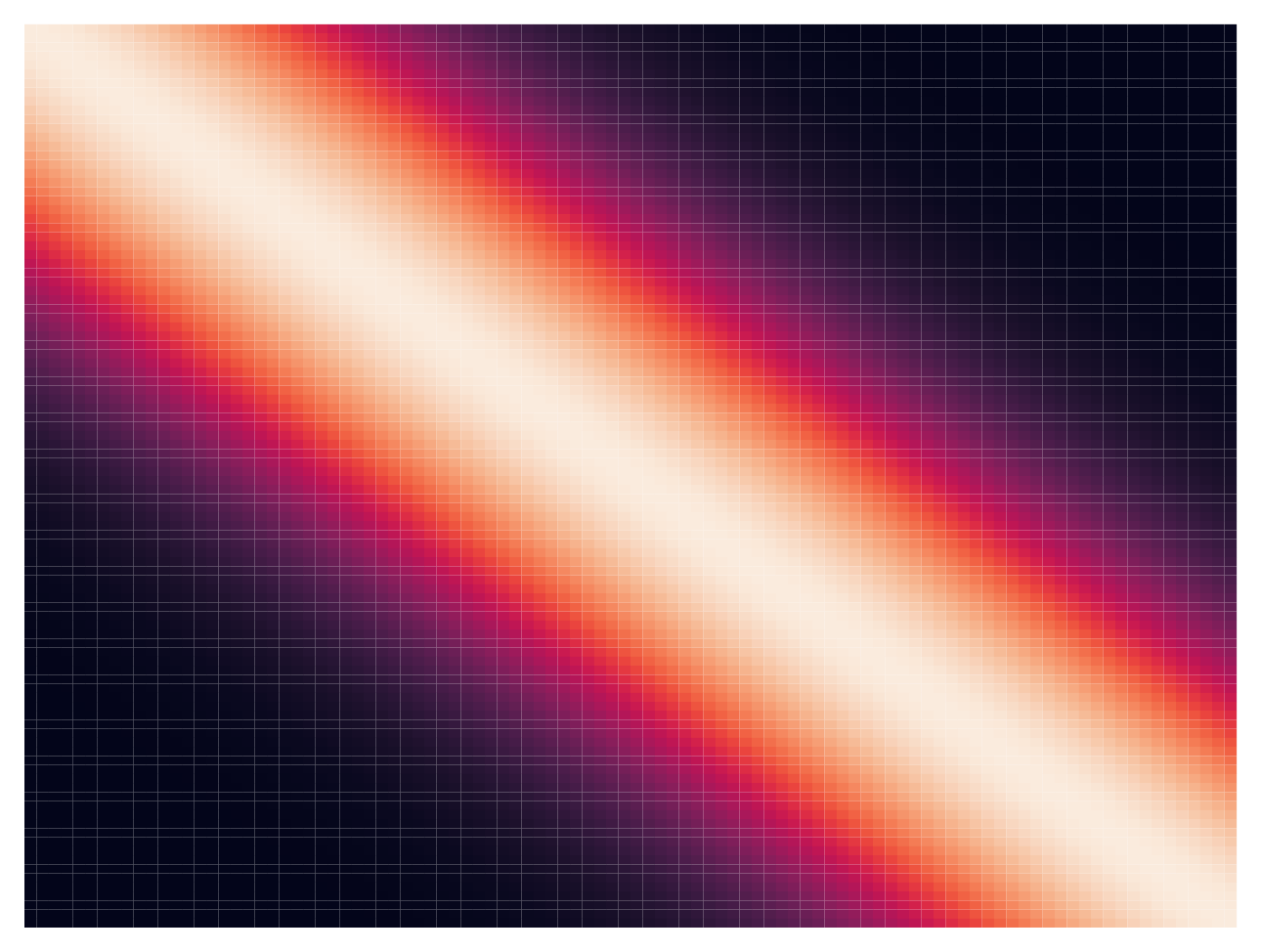}&
  \includegraphics[width=.22\linewidth]{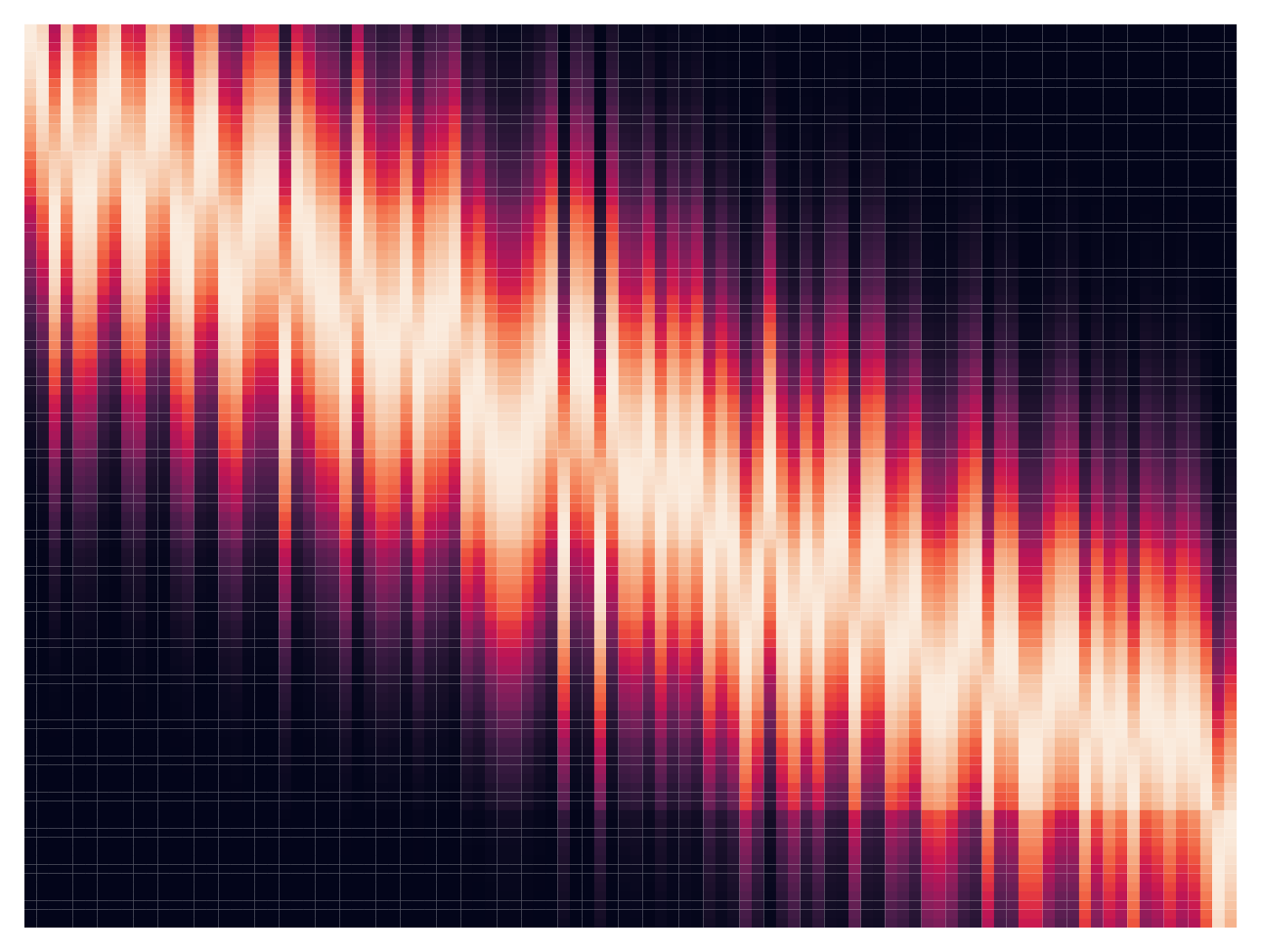}&
  \includegraphics[width=.22\linewidth]{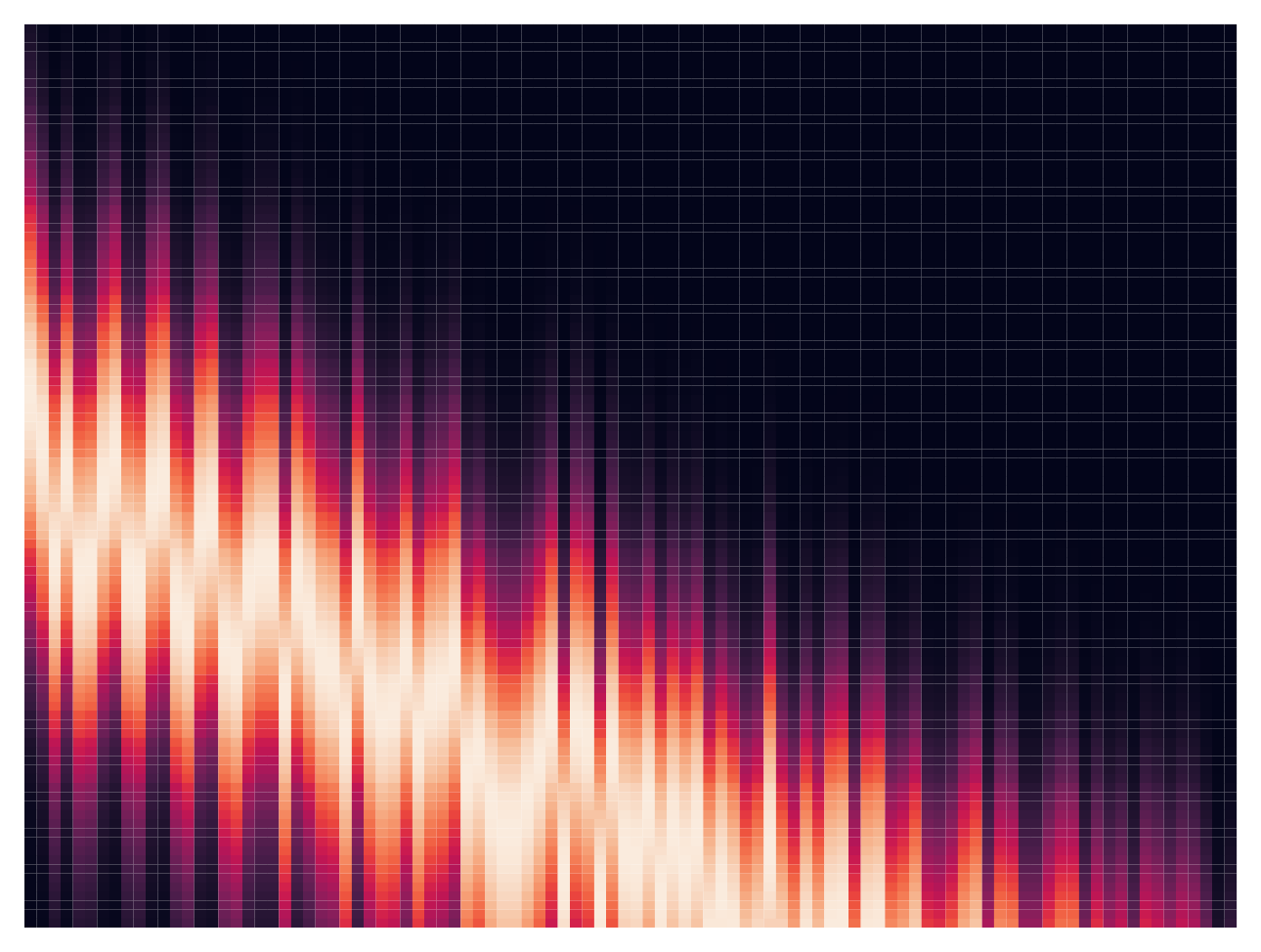}&
  \includegraphics[width=.22\linewidth]{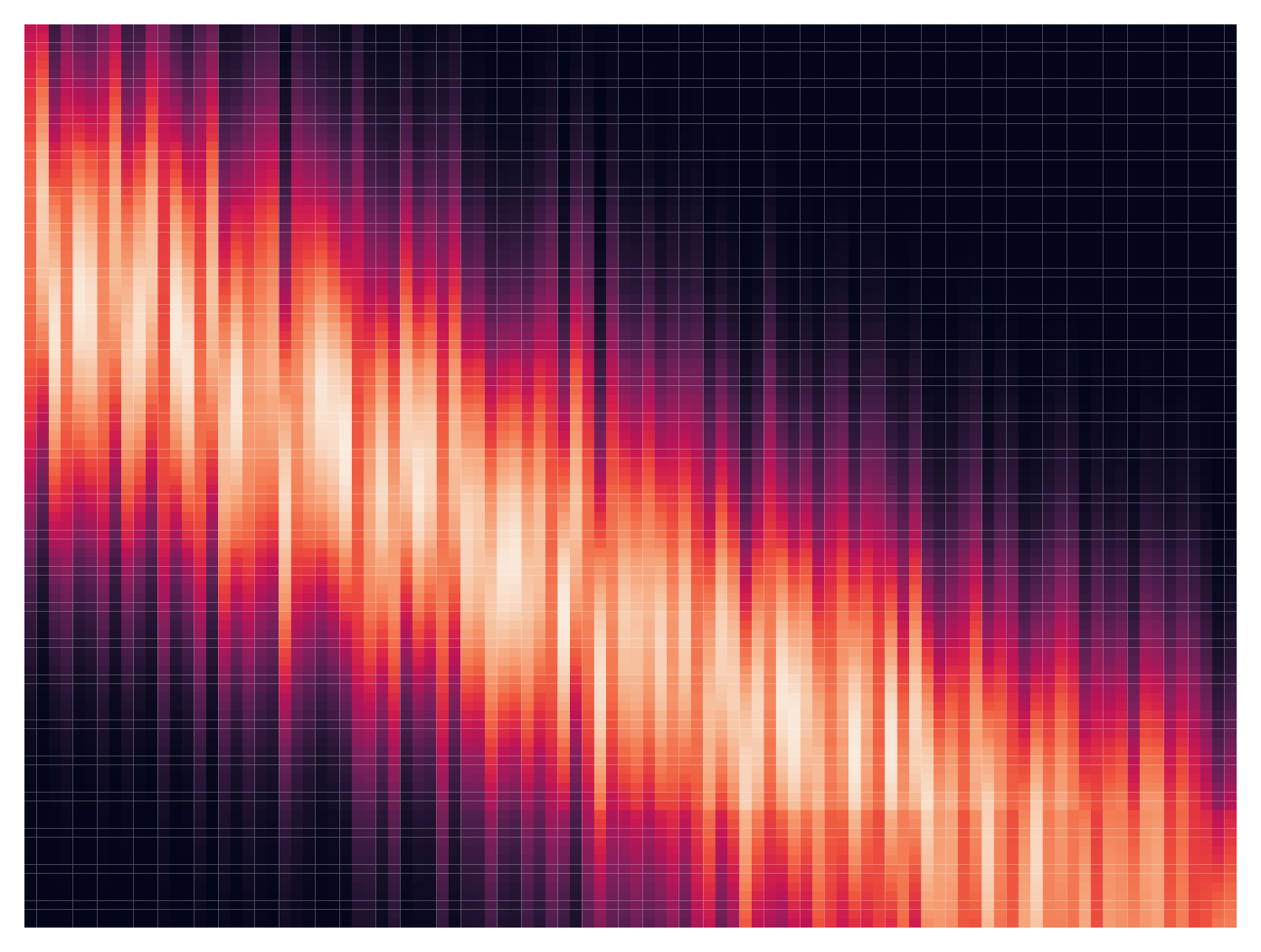}\\
  \small{\bf{(a)}}&
  \small{\bf{(b)}}&
  \small{\bf{(c)}}&
  \small{\bf{(d)}}
  \vspace{-3mm}
  \end{tabular}
    \caption{ {\bf Toy Example of Temporal Priors.} Light color denotes high alignment probability. (a) Consistency prior with the peak values appearing along the diagonal. (b) Ground-truth probability for which there exist many non-monotonic frames.  (c) Optimality prior from transportation matrix, with the peak values appearing on the locations of most similar pairs in the embedding space. (d) Gaussian mixture of (b) and (c), which more accurately represents the ground-truth, shown in (b), as compared to (a) or (c).
    \vspace{-4mm}
   }
    \label{fig:prior}
    \end{figure}

The first prior, which we call as \emph{Consistency Prior}, favors transportation of one sequence to the elements in the nearby temporal positions of the other sequence, and hence respects the overall temporal structure and the consistency in the order of the actions across sequences. With this prior, assignment matrix is very likely to have peak values along the diagonal and the values should gradually decrease along the direction perpendicular to the diagonal, as depicted by Fig.~\ref{fig:prior}(a). We can model this situation with a two-dimensional distribution, in which the distribution along any line perpendicular to the diagonal is a Gaussian distribution centered on the diagonal. We model the \emph{Consistency Prior} on the assignment matrix with a Gaussian as follows
\begin{align}
    \bm{P}_{c}(i,j) = \frac{1}{\sigma \sqrt{2\pi}}e^{-\frac{l^{2}_{c}(i,j)}{2\sigma^{2}}} \;,  \label{eq:homogeneous}   
\end{align}
where $l_{c}(i,j)$ is the distance from the position $(i,j)$ to the diagonal
\begin{align}
    l_{c}(i,j) = \frac{|i/N - j/M|}{\sqrt{1/N^{2}+1/M^{2}}}  \;.  \label{eq:l_homogeneous}
\end{align}
%

This prior, while modeling consistency across sequences, does not allow for handling unconstrained non-monotonic sequences. For example, in the extreme case of actions being performed in the exact reverse order across two sequences, the consistency prior would not be able to capture temporal variations. Similarly, for two sequences, in which there exists many non-monotonic frames, as depicted by Fig.~\ref{fig:prior}(b), this probability distribution would not ideally model the alignment. 

To be able to explicitly deal with non-monotonic sequences, we propose another prior, which we call, \emph{Optimality Prior}. Recall that the transport matrix $\bm{T}$ we compute in Eq.~\ref{eq:do} during the training process indicates the rough alignment between two video sequences and changes dynamically according to the temporal variations across sequences. We exploit this transport matrix to model another temporal prior. In particular, as depicted by Fig.~\ref{fig:prior}(c), we model our prior, such that the distribution along any line perpendicular to the diagonal is a Gaussian, centered at the intersection of the most likely alignment based on the transport matrix. We model the \emph{Optimality Prior} on the assignment matrix with
\begin{align}
    \bm{P}_{o}(i,j) = \frac{1}{\sigma \sqrt{2\pi}}e^{-\frac{l^{2}_{o}(i,j)}{2\sigma^{2}}} \;,  \label{eq:optimal}   
\end{align}
where $l_{o}(i,j)$ is the average distance from the position $(i,j)$, to the frame locations that give the optimal alignment, $(i,j_{o})$ and $(i_{o},j)$, given by the transport matrix
\begin{align}
    l_{o}(i,j) = \frac{|i/N - i_{o}/N|+|j/M - j_{o}/M|}{2\sqrt{1/N^{2}+1/M^{2}}}  \;.  \label{eq:l_optimal}
\end{align}

In short, Consistency Prior $\bm{P}_{c}$ represents the general case, in which the sequence pairs follow the same coarse ordering, while Optimality Prior $\bm{P}_{o}$ models the potential temporal variations across sequences. As shown by Fig.~\ref{fig:prior}(d), the ground truth distribution is more accurately represented by the combination of these two priors, which we formulate using a Gaussian Mixture Model, as follows:
\begin{align}
    \bm{P}(i,j) = \psi \bm{P_{c}}(i,j) + (1-\psi)\bm{P_{o}}(i,j) \;,  \label{eq:all}   
\end{align}
where $\psi \in [0,1]$ is a weighting parameter that we set to $1.0$ initially and decrease gradually over time to account for the fact that the learned transport matrix is less reliable in the very beginning of the training and becomes more robust in later stages. By enforcing temporal priors on optimal transport, our model is able to adaptively handle non-monotonic frames and temporal variations.

\begin{figure}[t]
\centering
\vspace{-2mm}
\includegraphics[width=0.9\linewidth]{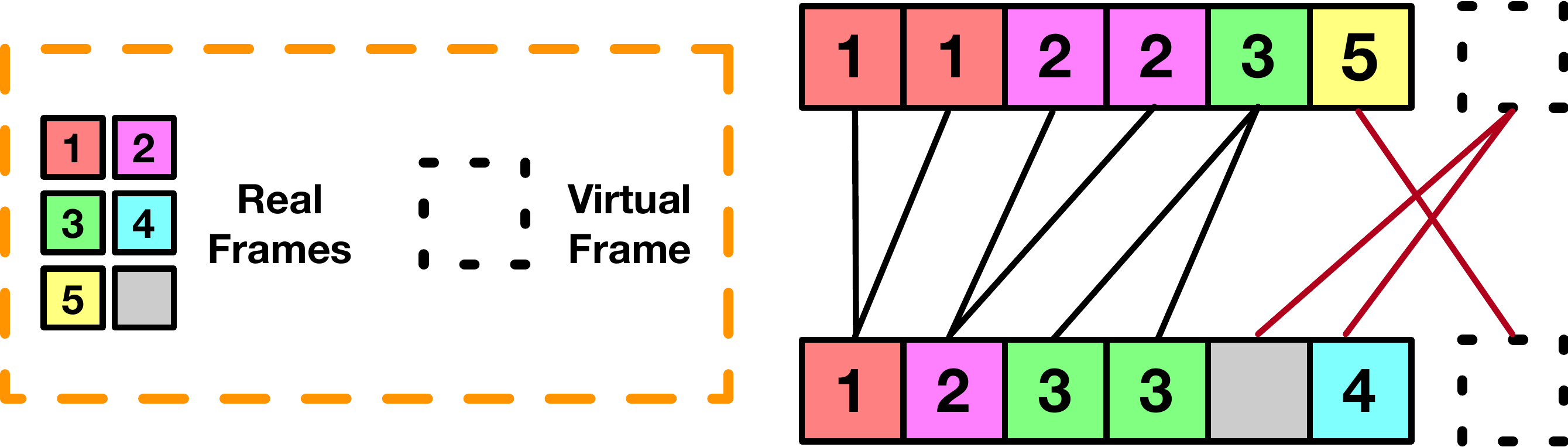}
\caption{\small {\bf Virtual Frame.} Virtual frame enables the model to handle  unmatched frames that should not be aligned. Redundant frames (shown with $4$ and $5$) and background frames (shown in gray) are explicitly assigned to it.}
\vspace{-4mm}
\label{fig:virtual}
\end{figure}

\subsection{Handling Background and Redundant Frames}
\label{sec:virtual}
\emph{Consistency} and \emph{Optimality} temporal priors enable our model to handle non-monotonic frames between video sequences. However they do not explicitly handle background and redundant frames, introduced in Sec.~\ref{sec:intro}. To be able to account for such frames in our model, we introduce an additional \emph{virtual frame} in the transport matrix so that unmatched frames are explicitly assigned to it, as shown in Fig.~\ref{fig:virtual}.

To this end, we augment the transport matrix, $\bm{T} \in \real^{N \times M}$, with an additional entry for each sequence to obtain  $\bm{\hat{T}} \in \real^{(N+1) \times (M+1)}$. The set of all feasible transportation matrices introduced in Eq.~\ref{eq:u} then becomes
\begin{equation}
     \resizebox{1.0\hsize}{!}{ $U(\bm{\hat{\alpha}},\bm{\hat{\beta}}):=\{\bm{\hat{T}} \in \real^{(N+1) \times (M+1)}    |\bm{\hat{T}}\bm{1}_{M+1}=\bm{\hat{\alpha}},\bm{\hat{T}}^{\top}\bm{1}_{N+1}=\bm{\hat{\beta}}\}$ } \nonumber
     \label{eq:ud} 
\end{equation}
where $\bm{\hat{\alpha}}$ and $\bm{\hat{\beta}}$ are the weight vectors expanded with one extra element to account for the virtual frame. If the chance of alignment with all the real frames is less than a certain threshold value, $\zeta$, we align this frame to the virtual frame instead. Note that many frames can be aligned with the virtual frame and the virtual frame does not follow the temporal priors we defined in Sec.~\ref{sec:temporal}.

\subsection{Training Loss}
\label{sec:loss}

\paragraph{VAVA Loss.} Our model accounts for temporal variations exhibited across real-world sequences with a differentiable formulation that relies on an optimal transport loss. We regularize our loss function by exploiting temporal priors, as explained in Sec.~\ref{sec:temporal}.
For the Consistency Prior described in Eq.~\ref{eq:homogeneous}, the large values of the transport matrix $\bm{\hat{T}}$ should be along the diagonal and the rest of the values should be small for other regions. Such a structure of the transport matrix can be measured with
\begin{align}
    I_{c}(\bm{\hat{T}}) = \sum_{i=1}^{N+1}\sum_{j=1}^{M+1}\frac{t_{ij}}{(\frac{i}{N+1}-\frac{j}{M+1})^{2}+1}  \;.  \label{eq:inverse_homogeneous}
\end{align}
where we add one more row and column for virtual frames as explained in Sec.~\ref{sec:virtual}, $I_{c}(\bm{\hat{T}})$ in Eq.~\ref{eq:inverse_homogeneous} is referred to as \emph{inverse difference moment} in literature~\cite{Albregtsen08,Su17} and will have large values for the region along the diagonal.

For the Optimality Prior described in Eq.~\ref{eq:optimal}, in which, large values appear in the most likely alignment locations given by the transport matrix, a similar structure can be captured with
\begin{small}
\begin{align}
    I_{o}(\bm{\hat{T}}) &= \sum_{i=1}^{N+1}\sum_{j=1}^{M+1} \frac{t_{ij}}{\frac{1}{2}d_{o}+1} \\
    d_{o} & = (\frac{i-i_{o}}{N+1})^{2}+(\frac{j-j_{o}}{M+1})^{2}  \nonumber
\end{align}
\end{small}
Our overall temporal prior that combines the Consistency Prior and the Optimality Prior can then be represented with the following loss function defined on the transport matrix
%
\begin{align}
    I(\bm{\hat{T}}) = \psi I_{c}(\bm{\hat{T}}) + (1-\psi)I_{o}(\bm{\hat{T}}) \;,  
    \label{eq:all_inverse}   
\end{align}
with the same $\psi$ as we defined in Eq.~\ref{eq:all}. For a smooth alignment, we further enforce the expected distribution to be similar to the temporal priors by minimizing the Kullback-Leibler(KL) divergence between the two matrices
\begin{align}
    KL(\bm{\hat{T}}||\bm{\hat{P}}) = \sum_{i=1}^{N+1}\sum_{j=1}^{M+1} t_{ij} \log\frac{t_{ij}}{p_{ij}} \;,  \label{eq:KL}   
\end{align}
where $\bm{\hat{P}}$ is as defined in Eq.~\ref{eq:all}, except that it is augmented with the virtual frame. Our variation-aware video alignment (VAVA) loss would therefore be defined by combining the temporal priors and the KL divergence within the optimal transport formulation (Eq.~\ref{eq:do}):
\begin{align}
  \hspace{-2mm}  L_{vava} = d_{O}(\X,\Y) - \lambda_{1}I(\bm{\hat{T}})+\lambda_2 KL(\bm{\hat{T}}||\bm{\hat{P}}) \;,  \label{eq:ot_loss}   
\end{align}
where $d_{O}(\X,\Y)$ is the Sinkhorn distance~\cite{Cuturi13}, as defined in Eq.~\ref{eq:do}, with extra row and column for virtual frames; $\lambda_1$ and $\lambda_2$ are hyper-parameters to weigh the two loss terms.

\paragraph{Contrastive Regularization.} Enforcing temporal priors on video alignment generally suffers from converging to trivial solutions~\cite{Haresh21,Su17}. The previous work~\cite{Haresh21} employs an {\it intra-video} contrastive loss term to regularize the training process. The {\it intra-video} contrastive loss for a given video embedding, $\X$, is defined as
\begin{multline}
    C(\X) = \sum_{i=1}^{N+1}\sum_{j=1}^{M+1} \mathds{1}_{|i-j|>\delta} \bm{W}(i,j)max(0,\lambda_{3} - \bm{\hat{D}}_{\X}(i,j)) \\
    +\mathds{1}_{|i-j|\leq \delta} \bm{W}(i,j)\bm{\hat{D}}_{\X}(i,j),  \label{eq:C_X}   
\end{multline}
where $\mathds{1}$ is an indicator function, which is $1$, if the condition is met, and, $0$ otherwise. $\bm{W}(i,j) = (i-j)^{2}+1$, is the distance in frame index and $\bm{D}_{\X}(i,j) = ||\x_{i}-\x_{j}||$, is the distance in the embedding space. $\delta$ is a window size for separating temporally far away and close frames and $\lambda_{3}$ is a margin parameter. This loss encourages close frames to be nearby in the embedding space, while penalizing temporally far away frames.

In our approach, in addition to using an intra-video contrastive loss term, we introduce an \emph{optimal transport guided inter-video contrastive loss} to regularize the training process. In particular, we propose to contrast video pairs based on their similarity given by optimal transport. As discussed in Sec.~\ref{sec:temporal}, our transport matrix provides an estimate of the alignment between two sequences in the training stage. We leverage this information to enforce an {\it inter-video contrastive loss}:
\begin{equation}
\hspace{-1mm} \resizebox{1.0\hsize}{!}{ $ C(\X,\Y) = \sum_{i=1}^{N+1}\sum_{j=1}^{M+1} - \mathds{1}_{{\bm {\bar{A}}}(i,j)}\bm{\hat{D}}_{\X,\Y}(i,j) 
    +\mathds{1}_{\bm{A}(i,j)} \bm{\hat{D}}_{\X,\Y}(i,j) $ }  
\label{eq:C_XY}   
\end{equation}
where ${\bm{A}}(i,j)$ denotes frames, $i$ and $j$, that yield the largest  $t_{i,j}$ value on our transport matrix for each row or column, while $\bar{\bm{A}}(i,j)$ denotes frames, $i$ and $j$, that are least likely for alignment, which have the smallest $t_{i,j}$ values. This loss encourages frames to have similar latent embeddings if they are expected to be aligned by our optimal transport formulation, and, if not, they are enforced to have dissimilar latent embeddings. Our total regularization term is defined by
\begin{align}
    L_{cr} = C(\X)+C(\Y)+C(\X,\Y)  \;,  \label{eq:r_loss}   
\end{align}

\paragraph{Final Loss.} Our final loss is obtained by combining the VAVA loss that enforces temporal priors on optimal transport (Eq.~\ref{eq:ot_loss}), along with contrastive regularization terms (Eq.~\ref{eq:r_loss}) that optimize for disentangled representations of frames within and across sequences.
\begin{align}
    L_{all} = L_{vava}+\gamma L_{cr}  \;,  \label{eq:all_loss}   
\end{align}
Here, $\gamma$ is a hyper-parameter to weigh the influence of the regularization term.

\newcommand{\vava}[0]{{\bf VAVA}}

\section{Evaluation}
\label{sec:experiments}

\paragraph{Datasets.}
We evaluate our approach on four different challenging datasets, namely COIN~\cite{Tang19}, IKEA ASM~\cite{Ben-Shabat20}, Pouring~\cite{Sermanet18}, and  Penn Action~\cite{Zhang13}.  COIN and IKEA ASM datasets exhibit large temporal variations and comprise of \emph{background frames}, \emph{redundant frames} and \emph{non-monotonic frames}, as described in Sec.~\ref{sec:intro}. We therefore use them to demonstrate the effectiveness of our approach in aligning sequential actions in unconstrained environments. Pouring and Penn Action datasets do not contain any such temporal variation, that is, action order is strictly monotonic and there are no background frames in the videos. We use these two datasets to benchmark our results against TCN~\cite{Sermanet18} and LAV~\cite{Haresh21}, which assume strict monotonic alignment. On the IKEA ASM dataset,~\cite{Haresh21} removes background frames for model training and evaluation. Since we aim to align unconstrained sequences, we instead keep the background frames by treating it as an additional category. In addition, in another evaluation setting, we remove the background frames to be able to compare against previous work~\cite{Haresh21}.
\vspace{-5mm}
\paragraph{Implementation Details.}
Following~\cite{Dwibedi19,Haresh21}, we use ResNet-50~\cite{He16} as the encoder network. The input videos are resized to $224 \times 224$. The  embeddings are extracted from the output of $Conv4c$ layer and are of size $14 \times 14 \times 1024$. We initialized our networks from ImageNet pre-trained models as in~\cite{Dwibedi19,Haresh21}. We set weighting of the regularization term, $\gamma$, in Eq.~\ref{eq:all_loss} as $0.5$.  We provide further details and ablation studies for the parameters we used in our Sup. Mat..
\vspace{-6mm}
\paragraph{Evaluation Metrics.}
Following~\cite{Dwibedi19,Haresh21}, we use three different metrics for our evaluation. We first train our encoder network on the training set without using any labels, and then evaluate the performance of our approach with the frozen embeddings. The first metric is {\it Phase Classification Accuracy}, which is the per frame classification accuracy for fine-grained action recognition. The second one is {\it Phase Progression(Progress)~\cite{Dwibedi19}}, which measures how well the {\it progress} of a process or action is captured by the embeddings. This metric assumes that actions are strictly consistent, thus is only suitable for monotonic datasets, that is Pouring and Penn Action, in our case. The last one is {\it Kendall's Tau~($\tau$)~\cite{Dwibedi19}}, which is a statistical measure that can determine how well-aligned two sequences are in time. Since this metric assumes strictly monotonic order of actions, it is only suitable for Pouring and Penn Action datasets. For all measures a higher score implies a better model.

\subsection{Comparison to the State-of-the-Art}
We evaluate the accuracy of our learned representation in the action phase classification task with an SVM classifier trained on a fraction $0.1$, $0.5$ and $1.0$ of the ground truth labels. We compare against the accuracy numbers reported in~\cite{Dwibedi19,Haresh21} on the Pouring, Penn Action and IKEA ASM datasets. Previous approaches do not report results on the unconstrained COIN dataset. Therefore we reproduce the results of these baselines on this dataset, to be able benchmark our results against them. To do so, we follow the implementation details of~\cite{Dwibedi19,Haresh21} and also validate the accuracy of our reproduced implementation on the Pouring, Penn Action and IKEA ASM datasets. We denote our Variation-Aware Video Alignment approach as \vava{} and report results on the  COIN,  IKEA ASM,  Pouring and Penn Action datasets in Table~\ref{tab:eva_merge}. 

	\begin{table}[t]
  \centering
  \scalebox{0.65}{
    \begin{tabular}{c|c|ccc|c|c}
      \toprule
 \multirow{2}{*}{{\bf Dataset}} &\multirow{2}{*}{{\bf Model}}  & \multicolumn{3}{c|}{{\bf Fraction of Labels}} & \multirow{2}{*}{{\bf Progress}} & \multirow{2}{*}{ \bm{$\tau$}} \\
 &  & {\bf 0.1} & {\bf 0.5} & {\bf 1.0} & & \\
  \midrule
 \multirow{8}{*}{{\bf COIN}}     &   Supervised Learning &  37.11 & 40.73  & 49.18 & - & -\\
    & Random Features & 29.50 & 30.29 & 30.38 & - & -\\
    & Imagenet Features & 31.32 & 34.74 & 37.43 & - & -\\
     \cline{2-7}
     & SAL~\cite{Misra16} & 34.69 &  39.23 &  40.32 & - & - \\
     &TCN~\cite{Sermanet18} & 34.87 & 39.73 &  40.51& - & -\\
    & TCC~\cite{Dwibedi19} & 35.87 &  39.56 &  40.66& - & -\\
     &LAV~\cite{Haresh21} & 36.79 & 38.85 & 39.81 & - & -\\
    & \vava{}(ours) & {\bf 43.77} & {\bf 46.18} & {\bf 47.26} & - & -\\
    
  \midrule
 \multirow{8}{*}{\vbox{\hbox{\strut {\bf IKEA ASM} } \hbox{\strut No Background}}}    &  Supervised Learning & 21.76 & 30.26 & 33.81 & - & -\\
     &Random Features & 17.89 & 17.89  & 17.89 & - & -\\
     &Imagenet Features & 18.05 & 19.27 & 19.50 & - & -\\
     \cline{2-7}
      & SAL~\cite{Misra16} & 21.68 & 21.72 & 22.14 & - & -\\
   &TCN~\cite{Sermanet18} & 25.17 & 25.70 & 26.80 & - & -\\
     &TCC~\cite{Dwibedi19} & 24.74 & 25.22 & 26.46 & - & -\\
     &LAV~\cite{Haresh21} & 29.78 & 29.85 & 30.43 & - & -\\
    &\vava{}(ours) & {\bf 31.66} & {\bf 33.79} & {\bf 32.91} & - & -\\

  \midrule
 \multirow{8}{*}{\vbox{\hbox{\strut {\bf IKEA ASM} } \hbox{\strut Background}}}   & Supervised Learning & 20.74 & 25.61 & 31.92 & - & -\\
     &Random Features & 17.03 & 17.41 &17.61 & - & -\\
     &Imagenet Features & 17.27 & 18.02 & 18.64 & - & -\\
     \cline{2-5}
     & SAL~\cite{Misra16} & 22.94 & 23.43 & 25.46 & - & -\\
     &TCN~\cite{Sermanet18} & 22.51 & 25.47 & 25.88 & - & -\\
     &TCC~\cite{Dwibedi19} & 22.70 & 25.04 & 25.63  & - & -\\
     &LAV~\cite{Haresh21} & 23.19 & 25.47 & 25.54 & - & -\\
    &\vava{}(ours) & {\bf 29.12}  & {\bf 29.95} & {\bf 29.10} & - & -\\
    
  \midrule
  
 \multirow{8}{*}{{\bf Pouring}}   & Supervised Learning & 75.43 & 86.14 & 91.55 & - & -\\
    & Random Features & 42.73 & 45.94 & 46.08 & - & -\\
    & Imagenet Features & 43.85 & 46.06 & 51.13 & - & -\\
     \cline{2-7}
    &  SAL~\cite{Misra16} & 85.68 & 87.84 & 88.02 & 0.7451 & 0.7331 \\
    & TCN~\cite{Sermanet18} & 89.19  & 90.39 & 90.35 & 0.8057 & 0.8669\\
    & TCC~\cite{Dwibedi19} & 89.23 & 91.43 & 91.82 & 0.8030 & 0.8516\\
    & LAV~\cite{Haresh21} & 91.61 & {\bf 92.82} & {\bf 92.84}& 0.8054 & 0.8561\\
    &\vava{}(ours) & {\bf 91.65} & 91.79 & 92.45& {\bf 0.8361} & {\bf 0.8755}\\
    
      \midrule
      
  \multirow{8}{*}{{\bf Penn Action}} &   Supervised Learning &  67.10 & 82.78 &86.05 & - & -\\
   &  Random Features &  44.18 & 46.19 & 46.81 & - & -\\
   &  Imagenet Features & 44.96 & 50.91 & 52.86 & - & -\\
     \cline{2-7}
    &  SAL~\cite{Misra16} & 74.87 & 78.26 & 79.96 & 0.5943 & 0.6336 \\
    & TCN~\cite{Sermanet18} & 81.99 & 83.67 & 84.04 & 0.6762 & 0.7328\\
    & TCC~\cite{Dwibedi19} & 81.26 & 83.35 & 84.45 & 0.6726 & 0.7353 \\
    & GTA~\cite{Hadji21} &- & - & - & - & 0.7829 \\
    & LAV~\cite{Haresh21} & 83.56 & 83.95 & 84.25 & 0.6613 & 0.8047\\
    &\vava{}(ours) & {\bf 83.89} & {\bf 84.23} & {\bf 84.48} &  {\bf 0.7091} & {\bf 0.8053}\\
  \bottomrule
    \end{tabular}
  }
\vspace{-2mm}
  \caption{{\bf Benchmark Evaluation.}}
  \label{tab:eva_merge}
  \vspace{-6mm}
\end{table}

Our model clearly outperforms earlier work on the COIN and IKEA ASM datasets which feature temporal variations that are exhibited by many real world applications. Particularly, the improvement over state-of-the-art methods is around $7\%$ (with a relative improvement of $20\%$) on the COIN dataset, which demonstrates the effectiveness of our approach for aligning sequential actions across unlabeled videos from in-the-wild settings. Similarly, \vava{} achieves $5\%$ improvement (with a relative increase of $25\%$) over existing approaches on the IKEA ASM dataset that shows the benefits of our approach in aligning videos that feature temporal variations.

For Pouring and Penn Action datasets that do not involve temporal variations, our approach still outperforms previous work in phase progression, Kendall's $\tau$ and most of the phase classification accuracies, which demonstrates the representation power of our framework in modeling the progress of actions and their temporal structure. Note also that Pouring dataset contains videos that follow a strict monotonic temporal order, and therefore methods that rely on the monotonicity assumption~\cite{Haresh21} are more likely to overfit to this dataset.
\begin{figure}[t]
\centering
\vspace{-3mm}
\includegraphics[width=0.8\linewidth]{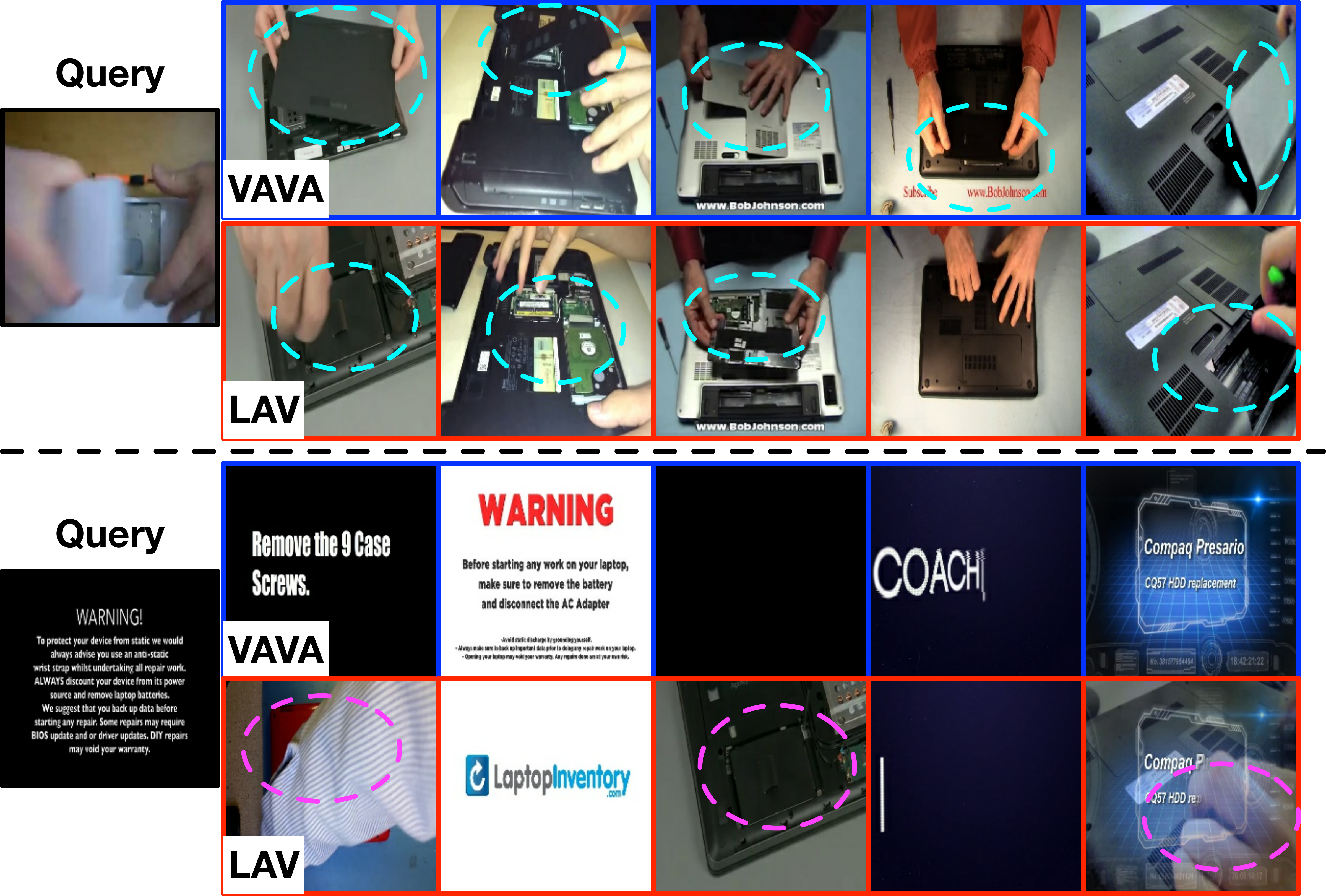}
\vspace{-2mm}
\caption{\small{\bf Frame Retrieval.} \hspace{-3mm} VAVA can precisely reason about fine grained actions and background frames. While we capture the fine-grained action of \emph{opening laptop cover},~\cite{Haresh21} retrieves images where laptop cover is already open (top). We recover background frames more consistently in comparison to~\cite{Haresh21} (bottom).}
\label{fig:retrieval}
\vspace{-2mm}
\end{figure}

	\begin{table}[t]
  \centering
  \scalebox{0.59}{
    \begin{tabular}{ccccccc|ccc}
      \toprule
 Intra- & Inter- & KL & Consistency  &  Optimality &  Virtual& Threshold  &\multicolumn{3}{c}{{\bf Fraction of Labels}}  \\
  Video & Video &  & Prior& Prior & Frame &  & {\bf 0.1} & {\bf 0.5} & {\bf 1.0}  \\
  \midrule
  \cmark &  &  & &  & & & 20.38  & 23.09 & 23.27  \\
  & \cmark &  & &  &  & & 19.46 & 22.58 &  22.94 \\
  \cmark & \cmark &  & &  & &  & 22.80  & 24.67 & 24.96  \\
 \cmark  & \cmark & \cmark &\cmark &  &  & &   24.75  & 27.35 &27.03 \\
 \cmark  & \cmark & \cmark& \cmark & \cmark &  &    & 27.81 & 28.03  & 28.59 \\
   & & \cmark & \cmark& \cmark & \cmark &  &   21.72& 22.47  & 23.60\\
  \cmark  &   \cmark & \cmark & \cmark & \cmark  &  & \cmark  &  26.49& 27.25 & 27.63 \\ 
  \cmark & \cmark & \cmark & &   & \cmark &   &  24.21 &  27.64 & 27.29 \\
    \cmark & \cmark & &\cmark & \cmark  & \cmark &  &   28.03 &  28.65 & 28.37 \\
  \cmark & \cmark & \cmark & \cmark & \cmark  & \cmark &   &  {\bf 29.12} & {\bf 29.95}  & {\bf 29.10} \\
  \bottomrule
    \end{tabular}
  }
\vspace{-2mm}
  \caption{{\bf Ablation.} We ablate each proposed term on IKEA ASM~\cite{Ben-Shabat20}. All proposed terms consistently improve performance.}
  \label{tab:ablation}
  \vspace{-6mm}
\end{table}

\subsection{Ablation Studies}

In Table~\ref{tab:ablation}, we provide an ablation study to demonstrate the influence of each design choice of~\vava{} on the accuracy of action phase classification.~\emph{Intra-Video}  and~\emph{Inter-Video} denote the effect of the contrastive loss terms we introduced in Eq.~\ref{eq:C_X} and Eq.~\ref{eq:C_XY} to regularize the training process.~\emph{KL} shows the effect of KL divergence regularization term. While~\emph{Consistency Prior} denotes the temporal prior, introduced in Eq.~\ref{eq:homogeneous},  that enforces time consistency across videos during alignment,~\emph{Optimality Prior} denotes the temporal prior introduced in Eq.~\ref{eq:optimal} that favors optimal matching of frames across videos.~\emph{Virtual Frame}  shows the effect of extra virtual frame we incorporated in the optimal transport formulation to address background and redundant frames. We further compare our~\emph{Virtual Frame} strategy to a~\emph{Threshold} approach, in which alignments with a low matching score are removed based on a tuned threshold.

As shown in Table.~\ref{tab:ablation}, all of our design choices consistently improve the accuracy of our algorithm.~\emph{Optimality Prior} tackles variations in the sequence order, whereas~\emph{Consistency Prior} allows for respecting the coarse-level temporal structure and consistency of videos. While they both individually improve the performance, the Gaussian Mixture Model that combines the two further boosts the accuracy, which demonstrates the complementary nature of each prior. We further demonstrate that~\emph{Virtual Frame} strategy significantly improves performance as compared to a model that does not include it and a model that uses a simpler thresholding based approach (\emph{Threshold}) to handle background frames. We also evaluate the influence of the \emph{Intra-Video} and \emph{Inter-Video} contrastive loss terms and demonstrate that they result in superior performance, by regularizing the self-supervised learning process. Besides, the KL divergence loss that encourages smooth alignment further improves the performance.  We present qualitative results of frame retrieval, in which we match the most similar frame with a given query frame, in Fig.~\ref{fig:retrieval}. As shown on this example, \vava{} is able to reliably align both regular action frames and background frames.

\begin{figure}[t]
\centering
\vspace{-7mm}
\includegraphics[width=0.92\linewidth]{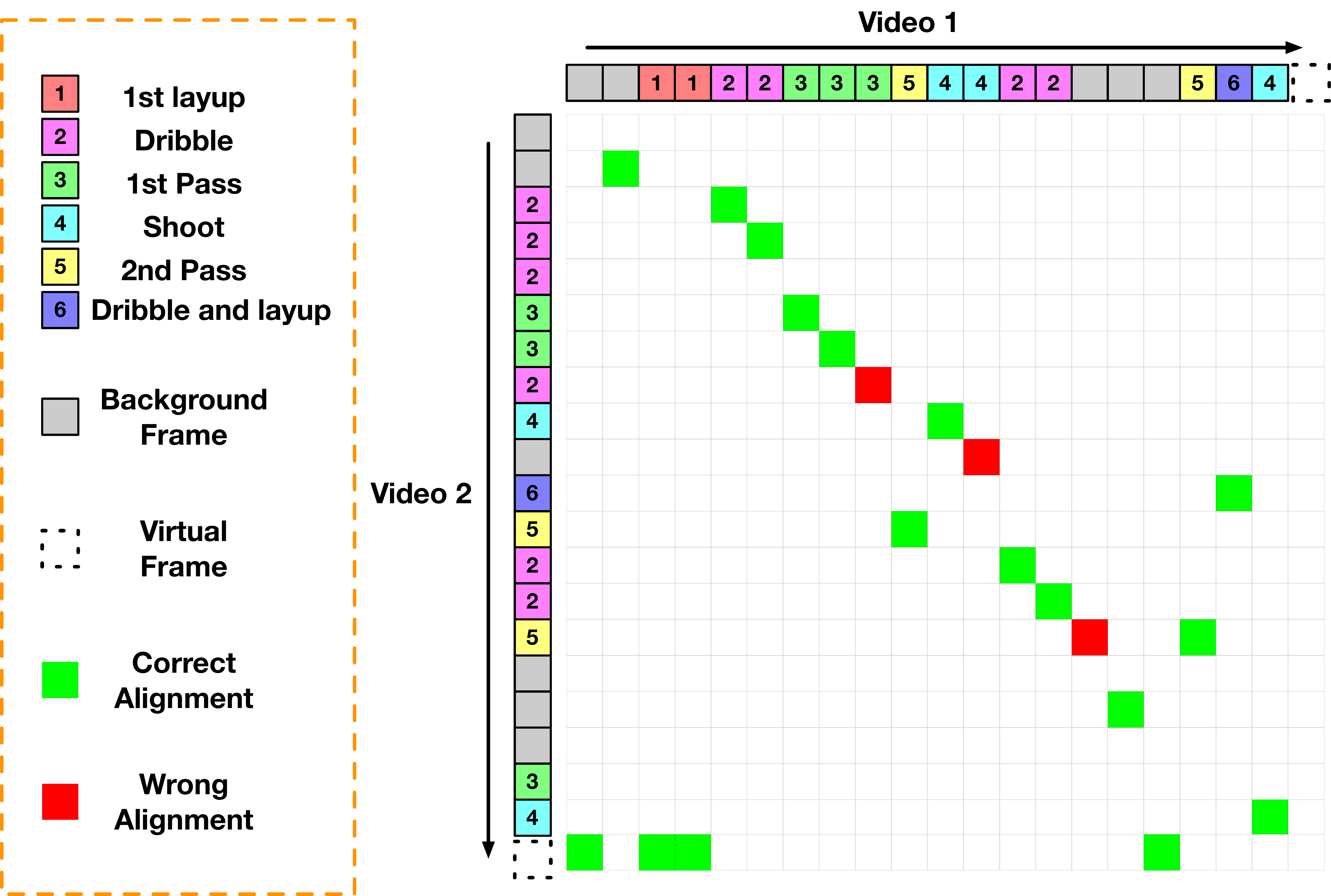}
\vspace{-2mm}
\caption{\small {\bf Example Alignment.} We align two videos from the \emph{Attend NBA Skills Challenge} task of the COIN dataset. For each frame in video $1$, we align it with the optimal match in the other sequence. Correct alignment means that the action frame is aligned with another frame with the same action. Redundant frames are aligned to the virtual frame and the background frames are aligned to either another background frame or the virtual frame. As can be seen by the high number of correct matches, our model can reliably align two sequences with temporal variations.}
\label{fig:alignment}
\vspace{-5mm}
\end{figure}

To demonstrate that our approach is able to align sequential actions in unconstrained environments, we visualize  the assignment matrix for a representative example on the COIN dataset, that feature different temporal variations involving \emph{background}, \emph{redundant} and \emph{non-monotonic} frames. As shown in Fig.~\ref{fig:alignment}, our model is able to align such sequences with high accuracy and brings in robustness against temporal variations, which makes it suitable for aligning sequential actions in-the-wild.

\vspace{-2mm}
\section{Conclusion}

In this paper, we propose a self-supervised learning framework that uses video alignment as a proxy task. The proposed VAVA approach is able to align sequential actions in-the-wild with an optimal transport based sequence alignment formulation. We further propose to enforce adaptive temporal priors on optimal transport, which efficiently handles temporal variations. Our experiments show that VAVA generally outperforms the state-of-the-art on the Pouring, Penn Action, IKEA ASM and COIN dataset. Our future work will explore applications of video alignment for AR-based task guidance and procedure learning.

{\bf Acknowledgments}  This work was completed during an internship at Microsoft Mixed Reality \& AI Lab. We would like to thank D. Dwibedi and S. Haresh for answering questions about TCC and LAV.


{\small
\bibliographystyle{ieee_fullname}
\bibliography{short,string,egbib}

\begin{thebibliography}{10}\itemsep=-1pt

\bibitem{Albregtsen08}
Fritz Albregtsen.
\newblock {Statistical texture measures computed from gray level coocurrence
  matrices}.
\newblock {\em Image processing laboratory, department of informatics,
  university of oslo}, 2008.

\bibitem{Anderson58}
Theodore~Wilbur Anderson.
\newblock {An introduction to multivariate statistical analysis}.
\newblock {\em Wiley New York}, 1958.

\bibitem{Andrew13}
Galen Andrew, Raman Arora, Jeff Bilmes, and Karen Livescu.
\newblock {Deep canonical correlation analysis}.
\newblock In {\em International Conference on Machine Learning}, 2015.

\bibitem{Behrmann21}
Nadine Behrmann, Mohsen Fayyaz, Juergen Gall, and Mehdi Noroozi.
\newblock {Long Short View Feature Decomposition via Contrastive Video
  Representation Learning}.
\newblock In {\em International Conference on Computer Vision}, 2021.

\bibitem{Ben-Shabat20}
Yizhak Ben-Shabat, Xin Yu, Fatemeh~Sadat Saleh, Dylan Campbell, Cristian
  Rodriguez-Opazo, Hongdong Li, and Stephen Gould.
\newblock {The IKEA ASM Dataset: Understanding People Assembling Furniture
  through Actions, Objects and Pose}.
\newblock In {\em arXiv Preprint}, 2020.

\bibitem{Bengio09}
Yoshua Bengio and James Bergstra.
\newblock {Slow, Decorrelated Features for Pretraining Complex Cell-Like
  Networks}.
\newblock In {\em Advances in Neural Information Processing Systems}, 2009.

\bibitem{Berndt94}
Donald~J. Berndt and James Clifford.
\newblock {Using dynamic time warping to find patterns in time series}.
\newblock In {\em KDD workshop}, 1994.

\bibitem{Carreira17}
Joao Carreira and Andrew Zisserman.
\newblock {Quo Vadis, Action Recognition? A New Model and the Kinetics
  Dataset.}
\newblock In {\em Conference on Computer Vision and Pattern Recognition}, 2017.

\bibitem{Chen20}
Ting Chen, Simon Kornblith, Mohammad Norouzi, and Geoffrey Hinton.
\newblock {A Simple Framework for Contrastive Learning of Visual
  Representations}.
\newblock In {\em International Conference on Learning Representations}, 2020.

\bibitem{Choi20}
Jinwoo Choi, Gaurav Sharma, Samuel Schulter, and Jia-Bin Huang.
\newblock {Shuffle and Attend: Video Domain Adaptation}.
\newblock In {\em European Conference on Computer Vision}, 2020.

\bibitem{Coskun21}
Huseyin Coskun, Zeeshan Zia, Bugra Tekin, Federica Bogo, Nassir Navab, Federico
  Tombari, and Harpreet Sawhney.
\newblock {Domain-Specific Priors and Meta Learning for Low-shot First-Person
  Action Recognition}.
\newblock {\em IEEE Transactions on Pattern Analysis and Machine Intelligence},
  2021.

\bibitem{Cuturi13}
Marco Cuturi.
\newblock {Sinkhorn distances: Lightspeed computation of optimal transport}.
\newblock In {\em Advances in Neural Information Processing Systems}, 2013.

\bibitem{Diba19}
Ali Diba, Vivek Sharma, Luc~Van Gool, and Rainer Stiefelhagen.
\newblock {Dynamonet: Dynamic Action and Motion Network}.
\newblock In {\em International Conference on Computer Vision}, 2019.

\bibitem{Diba21}
Ali Diba, Vivek Sharma, Reza Safdari, Dariush Lotfi, M.~Saquib Sarfraz, Rainer
  Stiefelhagen, and Luc~Van Gool.
\newblock {Vi2CLR: Video and Image for Visual Contrastive Learning of
  Representation}.
\newblock In {\em International Conference on Computer Vision}, 2021.

\bibitem{Dogan18}
Pelin Dogan, Boyang Li, Leonid Sigal, and Markus Gross.
\newblock {A Neural Multi-sequence Alignment TeCHnique (NeuMATCH)}.
\newblock In {\em Conference on Computer Vision and Pattern Recognition}, 2018.

\bibitem{Dwibedi19}
Debidatta Dwibedi, Yusuf Aytar, Jonathan Tompson, Pierre Sermanet, and Andrew
  Zisserman.
\newblock {Temporal Cycle-Consistency Learning}.
\newblock In {\em Conference on Computer Vision and Pattern Recognition}, 2019.

\bibitem{Dwibedi21}
Debidatta Dwibedi, Yusuf Aytar, Jonathan Tompson, Pierre Sermanet, and Andrew
  Zisserman.
\newblock {With a Little Help from My Friends: Nearest-Neighbor Contrastive
  Learning of Visual Representations}.
\newblock In {\em International Conference on Computer Vision}, 2021.

\bibitem{Feichtenhofer21}
Christoph Feichtenhofer, Haoqi Fan, Bo Xiong, Ross Girshick, and Kaiming He.
\newblock {A Large-Scale Study on Unsupervised Spatiotemporal Representation
  Learning}.
\newblock In {\em Conference on Computer Vision and Pattern Recognition}, 2021.

\bibitem{Feng19}
Zeyu Feng, Chang Xu, and Dacheng Tao.
\newblock {Self-Supervised Representation Learning by Rotation Feature
  Decoupling}.
\newblock In {\em Conference on Computer Vision and Pattern Recognition}, 2019.

\bibitem{Fernando17}
Basura Fernando, Hakan Bilen, Efstratios Gavves, and Stephen Gould.
\newblock { Self-Supervised Video Representation Learning with Odd-One-Out
  Networks}.
\newblock In {\em Conference on Computer Vision and Pattern Recognition}, 2017.

\bibitem{Gammulle19}
Harshala Gammulle, Simon Denman, Sridha Sridharan, and Clinton Fookes.
\newblock {Predicting the Future: A Jointly Learnt Model for Action
  Anticipation}.
\newblock In {\em International Conference on Computer Vision}, 2019.

\bibitem{Ge21}
Zheng Ge, Songtao Liu, Zeming Li, Osamu Yoshie, and Jian Sun.
\newblock {OTA: Optimal Transport Assignment for Object Detection}.
\newblock In {\em Conference on Computer Vision and Pattern Recognition}, 2021.

\bibitem{Gidaris18}
Spyros Gidaris, Praveer Singh, and Nikos Komodakis.
\newblock {Unsupervised Representation Learning by Predicting Image Rotations}.
\newblock In {\em International Conference on Learning Representations}, 2018.

\bibitem{Goroshin15}
Ross Goroshin, Joan Bruna, Jonathan Tompson, David Eigen, and Yann LeCun.
\newblock {Unsupervised Learning of Spatiotemporally Coherent Metrics}.
\newblock In {\em International Conference on Computer Vision}, 2015.

\bibitem{Hadji21}
Isma Hadji, Konstantinos~G. Derpanis, and Allan~D. Jepson.
\newblock {Representation Learning via Global Temporal Alignment and
  Cycle-Consistency}.
\newblock In {\em Conference on Computer Vision and Pattern Recognition}, 2021.

\bibitem{Hadsell06}
Raia Hadsell, Sumit Chopra, and Yann LeCun.
\newblock {Dimensionality Reduction by Learning an Invariant Mapping}.
\newblock In {\em Conference on Computer Vision and Pattern Recognition}, 2006.

\bibitem{Han19}
Tengda Han, Weidi Xie, and Andrew Zisserman.
\newblock {Video Representation Learning by Dense Predictive Coding}.
\newblock In {\em International Conference on Computer Vision Workshops}, 2019.

\bibitem{Haresh21}
Sanjay Haresh, Sateesh Kumar, Huseyin Coskun, Shahram~Najam Syed, Andrey Konin,
  Muhammad~Zeeshan Zia, and Quoc-Huy Tran.
\newblock {Learning by Aligning Videos in Time}.
\newblock In {\em Conference on Computer Vision and Pattern Recognition}, 2021.

\bibitem{He20}
Kaiming He, Haoqi Fan, Yuxin Wu, Saining Xie, and Ross Girshick.
\newblock {Momentum Contrast for Unsupervised Visual Representation Learning}.
\newblock In {\em Conference on Computer Vision and Pattern Recognition}, 2020.

\bibitem{He16}
Kaiming He, Xiangyu Zhang, Shaoqing Ren, and Jian Sun.
\newblock {Deep Residual Learning for Image Recognition}.
\newblock In {\em Conference on Computer Vision and Pattern Recognition}, 2016.

\bibitem{Hinton94}
Geoffrey~E. Hinton and Richard~S. Zemel.
\newblock {Autoencoders, Minimum Description Length and Helmholtz Free Energy}.
\newblock In {\em Advances in Neural Information Processing Systems}, 1994.

\bibitem{Kotar21}
Klemen Kotar, Gabriel Ilharco, Ludwig Schmidt, Kiana Ehsani, and Roozbeh
  Mottaghi.
\newblock {Contrasting Contrastive Self-Supervised Representation Learning
  Pipelines}.
\newblock In {\em International Conference on Computer Vision}, 2021.

\bibitem{Kwon21}
Taein Kwon, Bugra Tekin, Jan Stuhmer, Federica Bogo, and Marc Pollefeys.
\newblock {H2O: Two Hands Manipulating Objects for First Person Interaction
  Recognition}.
\newblock In {\em International Conference on Computer Vision}, 2021.

\bibitem{Larsson16}
Gustav Larsson, Michael Maire, and Gregory Shakhnarovich.
\newblock {Learning Representations for Automatic Colorization}.
\newblock In {\em European Conference on Computer Vision}, 2016.

\bibitem{Larsson17}
Gustav Larsson, Michael Maire, and Gregory Shakhnarovich.
\newblock {Colorization as a Proxy Task for Visual Understanding}.
\newblock In {\em Conference on Computer Vision and Pattern Recognition}, 2017.

\bibitem{Lee17}
Hsin-Ying Lee, Jia-Bin Huang, Maneesh Singh, and Ming-Hsuan Yang.
\newblock {Unsupervised Representation Learning by Sorting Sequences}.
\newblock In {\em International Conference on Computer Vision}, 2017.

\bibitem{Li21}
Ruibo Li, Guosheng Lin, and Lihua Xie.
\newblock {Self-Point-Flow: Self-Supervised Scene Flow Estimation from Point
  Clouds with Optimal Transport and Random Walk}.
\newblock In {\em Conference on Computer Vision and Pattern Recognition}, 2021.

\bibitem{Lin21}
Yuanze Lin, Xun Guo, and Yan Lu.
\newblock {Self-Supervised Video Representation Learning with Meta-Contrastive
  Network}.
\newblock In {\em International Conference on Computer Vision}, 2021.

\bibitem{Liu21}
Weizhe Liu, David Ferstl, Samuel Schulter, Lukas Zebedin, Pascal Fua, and
  Christian Leistner.
\newblock {Domain Adaptation for Semantic Segmentation via Patch-Wise
  Contrastive Learning}.
\newblock In {\em arXiv Preprint}, 2021.

\bibitem{Misra16}
Ishan Misra, C.~Lawrence Zitnick, and Martial Hebert.
\newblock {Shuffle and Learn: Unsupervised Learning Using Temporal Order
  Verification}.
\newblock In {\em European Conference on Computer Vision}, 2016.

\bibitem{Mobahi09}
Hossein Mobahi, Ronan Collobert, and Jason Weston.
\newblock {Deep Learning from Temporal Coherence in Video}.
\newblock In {\em International Conference on Machine Learning}, 2009.

\bibitem{Noroozi17}
Mehdi Noroozi, Hamed Pirsiavash, and Paolo Favaro.
\newblock {Representation Learning by Learning to Count}.
\newblock In {\em International Conference on Computer Vision}, 2017.

\bibitem{Pan21}
Tian Pan, Yibing Song, Tianyu Yang, Wenhao Jiang, and Wei Liu.
\newblock {VideoMoCo: Contrastive Video Representation Learning with Temporally
  Adversarial Examples}.
\newblock In {\em Conference on Computer Vision and Pattern Recognition}, 2021.

\bibitem{Puy20}
Gilles Puy, Alexandre Boulch, and Renaud Marlet.
\newblock {FLOT: Scene Flow on Point Clouds guided by Optimal Transport}.
\newblock In {\em European Conference on Computer Vision}, 2020.

\bibitem{Qian21}
Rui Qian, Tianjian Meng, Boqing Gong, Ming-Hsuan Yang, Huisheng Wang, Serge
  Belongie, and Yin Cui.
\newblock {Spatiotemporal Contrastive Video Representation Learning}.
\newblock In {\em Conference on Computer Vision and Pattern Recognition}, 2021.

\bibitem{Sarlin20}
Paul-Edouard Sarlin, Daniel DeTone, Tomasz Malisiewicz, and Andrew Rabinovich.
\newblock {SuperGlue: Learning Feature Matching with Graph Neural Networks}.
\newblock In {\em Conference on Computer Vision and Pattern Recognition}, 2020.

\bibitem{Sermanet18}
Pierre Sermanet, Corey Lynch, Yevgen Chebotar, Jasmine Hsu, Eric Jang, Stefan
  Schaal, and Sergey Levine.
\newblock {Time-Contrastive Networks: Self-Supervised Learning from Video}.
\newblock In {\em International Conference on Robotics and Automation}, 2018.

\bibitem{Serrurier21}
Mathieu Serrurier, Franck Mamalet, Alberto González-Sanz, Thibaut Boissin,
  Jean-Michel Loubes, and Eustasio del Barrio.
\newblock {Achieving robustness in classification using optimal transport with
  hinge regularization}.
\newblock In {\em Conference on Computer Vision and Pattern Recognition}, 2021.

\bibitem{Srivastava15}
Nitish Srivastava, Elman Mansimov, and Ruslan Salakhutdinov.
\newblock {Unsupervised Learning of Video Representations Using LSTMs}.
\newblock In {\em International Conference on Machine Learning}, 2015.

\bibitem{Su17}
Bing Su and Gang Hua.
\newblock {Order-preserving Wasserstein Distance for Sequence Matching}.
\newblock In {\em Conference on Computer Vision and Pattern Recognition}, 2017.

\bibitem{Tang19}
Yansong Tang, Dajun Ding, Yongming Rao, Yu Zheng, Danyang Zhang, Lili Zhao,
  Jiwen Lu, and Jie Zhou.
\newblock {COIN: A Large-scale Dataset for Comprehensive Instructional Video
  Analysis }.
\newblock In {\em Conference on Computer Vision and Pattern Recognition}, 2019.

\bibitem{Tran15}
Du Tran, Lubomir Bourdev, Rob Fergus, Lorenzo Torresani, and Manohar Paluri.
\newblock {Learning Spatiotemporal Features with 3d Convolutional Networks}.
\newblock In {\em Conference on Computer Vision and Pattern Recognition}, 2018.

\bibitem{Tran18}
Du Tran, Heng Wang, Lorenzo Torresani, Jamie Ray, Yann LeCun, and Manohar
  Paluri.
\newblock {A Closer Look at Spatiotemporal Convolutions for Action
  Recognition}.
\newblock In {\em Conference on Computer Vision and Pattern Recognition}, 2018.

\bibitem{Vondrick16}
Carl Vondrick, Hamed Pirsiavash, and Antonio Torralba.
\newblock {Generating Videos with Scene Dynamics}.
\newblock In {\em Advances in Neural Information Processing Systems}, 2016.

\bibitem{Wang18}
Xiaolong Wang, Ross Girshick, Abhinav Gupta, and Kaiming He.
\newblock {Non-local neural networks}.
\newblock In {\em Conference on Computer Vision and Pattern Recognition}, 2018.

\bibitem{Wei18}
Donglai Wei, Joseph Lim, Andrew Zisserman, and William~T. Freeman.
\newblock { Learning and Using the Arrow of Time}.
\newblock In {\em Conference on Computer Vision and Pattern Recognition}, 2018.

\bibitem{Xu19}
Dejing Xu, Jun Xiao, Zhou Zhao, Jian Shao, Di Xie, and Yueting Zhuang.
\newblock {Self-Supervised Spatiotemporal Learning via Video Clip Order
  Prediction}.
\newblock In {\em Conference on Computer Vision and Pattern Recognition}, 2019.

\bibitem{Xu20}
Renjun Xu, Pelen Liu, Liyan Wang, Chao Chen, and Jindong Wang.
\newblock {Reliable Weighted Optimal Transport for Unsupervised Domain
  Adaptation}.
\newblock In {\em Conference on Computer Vision and Pattern Recognition}, 2020.

\bibitem{Yuan21}
Xin Yuan, Zhe Lin, Jason Kuen, Jianming Zhang, Yilin Wang, Michael Maire,
  Ajinkya Kale, and Baldo Faieta.
\newblock {Multimodal Contrastive Training for Visual Representation Learning}.
\newblock In {\em Conference on Computer Vision and Pattern Recognition}, 2021.

\bibitem{Zhang13}
Weiyu Zhang, Menglong Zhu, and Konstantinos~G. Derpanis.
\newblock {From actemes to action: A strongly-supervised representation for
  detailed action understanding}.
\newblock In {\em International Conference on Computer Vision}, 2013.

\bibitem{Zheng21}
Mingkai Zheng, Fei Wang, Shan You, Chen Qian, Changshui Zhang, Xiaogang Wang,
  and Chang Xu.
\newblock {Weakly Supervised Contrastive Learning}.
\newblock In {\em International Conference on Computer Vision}, 2021.

\bibitem{Zou12}
Will Zou, Shenghuo Zhu, Kai Yu, and Andrew~Y. Ng.
\newblock {Deep Learning of Invariant Features via Simulated Fixations in
  Video}.
\newblock In {\em Advances in Neural Information Processing Systems}, 2012.

\bibitem{Zou11}
Will~Y. Zou, Andrew~Y. Ng, and Kai Yu.
\newblock {Unsupervised Learning of Visual Invariance with Temporal Coherence}.
\newblock In {\em Advances in Neural Information Processing Systems Workshops},
  2012.

\end{thebibliography}
}

\end{document}